\documentclass[final]{cvpr} 
\usepackage{times}
\usepackage{epsfig}
\usepackage{graphicx}
\usepackage{amsmath}
\usepackage{amssymb}

\usepackage[square,numbers,sort&compress]{natbib}
\usepackage{enumitem}
\usepackage{algpseudocode}
\usepackage[dvipsnames]{xcolor}
\usepackage[font=small,labelfont=bf]{caption}
\usepackage{array}
\usepackage{multirow}
\usepackage{booktabs}
\usepackage{algorithm}
\usepackage{subcaption}
\usepackage[normalem]{ulem}
\usepackage{xparse}
\usepackage{pifont}
\usepackage{bm}
\usepackage{threeparttable}

\usepackage{listings}
\usepackage{mwe} \usepackage{makecell}
\usepackage{color, colortbl}

\usepackage[scaled=0.85]{DejaVuSansMono}

\usepackage[breaklinks=true,bookmarks=false,colorlinks,bookmarks=false, citecolor=ForestGreen]{hyperref}
\usepackage{cleveref}

\crefname{section}{\S}{\S\S}
\crefname{subsection}{\S}{\S\S}
\crefformat{table}{Table~#2#1#3}
\crefformat{figure}{Fig~#2#1#3}
\crefformat{equation}{Eq~#2#1#3}

\newcommand{\algorithmName}{DepthContrast\xspace}

\newcommand{\pointnetplus}{PointNet++\xspace}
\newcommand{\mink}{UNet\xspace}

\newcommand{\calD}{\mathcal{D}}
\newcommand{\calT}{\mathcal{T}}

\newcommand{\tx}{\widetilde{X}}
\newcommand{\bx}{\mathbf{X}}
\newcommand{\btx}{\mathbf{\tx}}
\newcommand{\bv}{\mathbf{v}}

\newcommand{\scanVideo}{ScanNet-vid\xspace}

\newcommand{\redVideo}{Redwood-vid\xspace}

\newcommand{\sunrgbd}{SUNRGBD\xspace}
\newcommand{\scannet}{ScanNet\xspace}
\newcommand{\modelnet}{ModelNet\xspace}
\newcommand{\stanford}{S3DIS\xspace}
\newcommand{\synthia}{Synthia\xspace}
\newcommand{\matterport}{Matterport3D\xspace}

\newlength\savewidth\newcommand\shline{\noalign{\global\savewidth\arrayrulewidth
  \global\arrayrulewidth 1pt}\hline\noalign{\global\arrayrulewidth\savewidth}}

\newlength\thinwidth

\definecolor{Gray}{gray}{0.92}
\newcolumntype{a}{>{\columncolor{Gray}}c}
\definecolor{LightCyan}{rgb}{0.88,1,1}
\definecolor{altRowColor}{gray}{0.92}
\definecolor{highlightRowColor}{rgb}{0.95, 0.95, 1}

\newcommand{\tabred}[1]{{\color{red} \scriptsize{#1}}}
\newcommand{\tabgreen}[1]{{\color{ForestGreen} \scriptsize{#1}}}

\def\confYear{CVPR 2021}

\begin{document}

\title{Self-Supervised Pretraining of 3D Features on \emph{any} Point-Cloud}

\author{
Zaiwei Zhang$^{1,2}$\thanks{Work done during an internship at Facebook.} \quad \quad Rohit Girdhar$^1$ \quad \quad Armand Joulin$^1$ \quad \quad Ishan Misra$^1$ \\
{\normalsize $^1$Facebook AI Research \quad \quad $^2$The University of Texas at Austin }
}

\maketitle

\begin{abstract}
Pretraining on large labeled datasets is a prerequisite to achieve good performance in many computer vision tasks like 2D object recognition, video classification etc.
However, pretraining is not widely used for 3D recognition tasks where state-of-the-art methods train models from scratch.
A primary reason is the lack of large annotated datasets because 3D data is both difficult to acquire and time consuming to label.
We present a simple self-supervised pretraining method that can work with any 3D data --- single or multi-view, indoor or outdoor, acquired by varied sensors, without 3D registration.
We pretrain standard point cloud and voxel based model architectures, and show that joint pretraining further improves performance.
We evaluate our models on 9 benchmarks for object detection, semantic segmentation, and object classification, where they achieve state-of-the-art results and can outperform supervised pretraining.
We set a new state-of-the-art for object detection on ScanNet (69.0\% mAP) and SUNRGBD (63.5\% mAP).
Our pretrained models are label efficient and improve performance for classes with few examples.
Code and models are available at \url{https://github.com/facebookresearch/DepthContrast}
\end{abstract}

\section{Introduction}

Pretraining visual features on large labeled datasets is a pre-requisite to achieve good performance when access to annotations is limited~\cite{mahajan2018exploring,ghadiyaram2019large,sun2017revisiting,kolesnikov2019big}.
More recently, self-supervised pretraining has become a popular alternative to supervised pretraining especially for tasks where annotations are time-consuming, such as detection and segmentation in images~\cite{he2019momentum,caron2020swav,henaff2019data,tian2019contrastive,misra2019self} or tracking in videos~\cite{jabri2020space}.
In 3D computer vision, single-view depth scans are easy to acquire while reconstructed 3D scenes and annotations are difficult to obtain.
Reconstructing a 3D scene requires registering and aligning multiple depth maps captured in a static environment, which may fail easily if there exists fast camera motion or odometry drift~\cite{Choi2016}. The resulting 3D scene is a point cloud composed of thousands of 3D points that need to be annotated individually in the case of segmentation or by groups for detection, which is time-consuming and laborious. For example, it takes around 22 minutes to annotate a single scene in ScanNet~\cite{Dai_2017_CVPR_scannet}.
Thus, 3D data acquisition and annotation both require significantly more effort than images or videos, and there is no existing weak supervision, like image tags, to shortcut the process~\cite{joulin2016learning,mahajan2018exploring,sun2017revisiting}.
This cumbersome annotation process results in a lack of large annotated 3D datasets. However, consumer-grade depth sensors have become more easily accessible and simpler to use, \eg, in phones~\cite{franklin2020ipad,stein2020iphone,samsung-depth-sensor}, leading to large quantities of single-view raw depth maps.
These depth maps can be leveraged to pretrain 3D features, but, there is surprisingly little work that can be applied.

\begin{figure}[!t]
	\centering
			\includegraphics[height=2in]{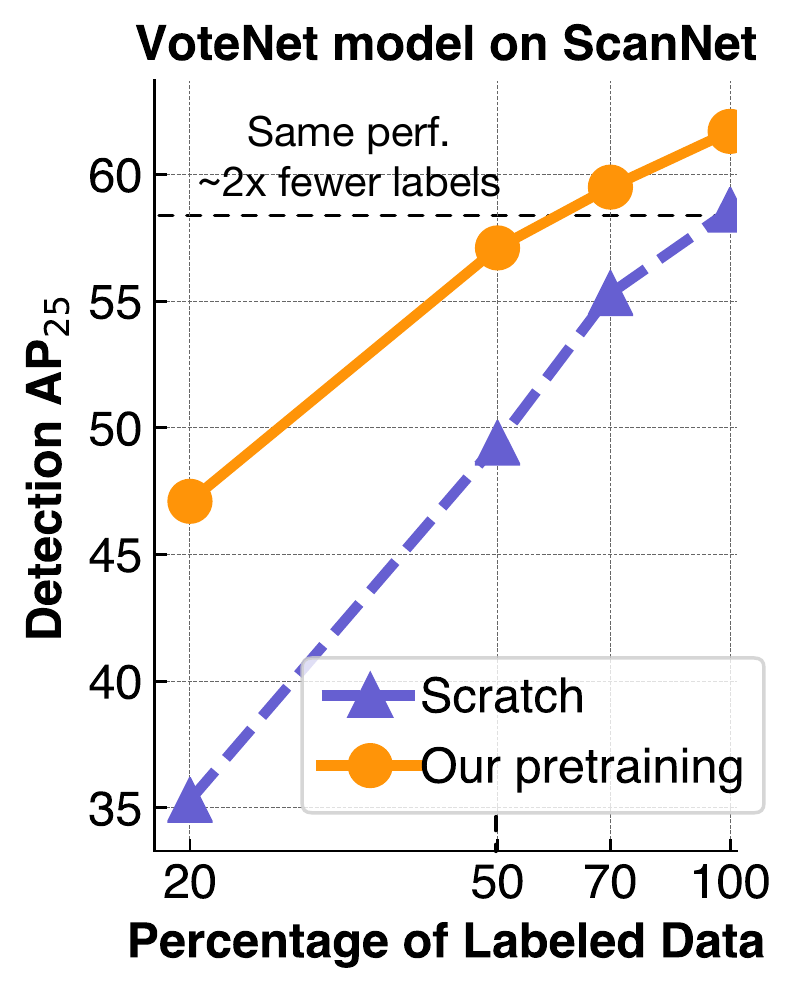}
	\includegraphics[height=2in]{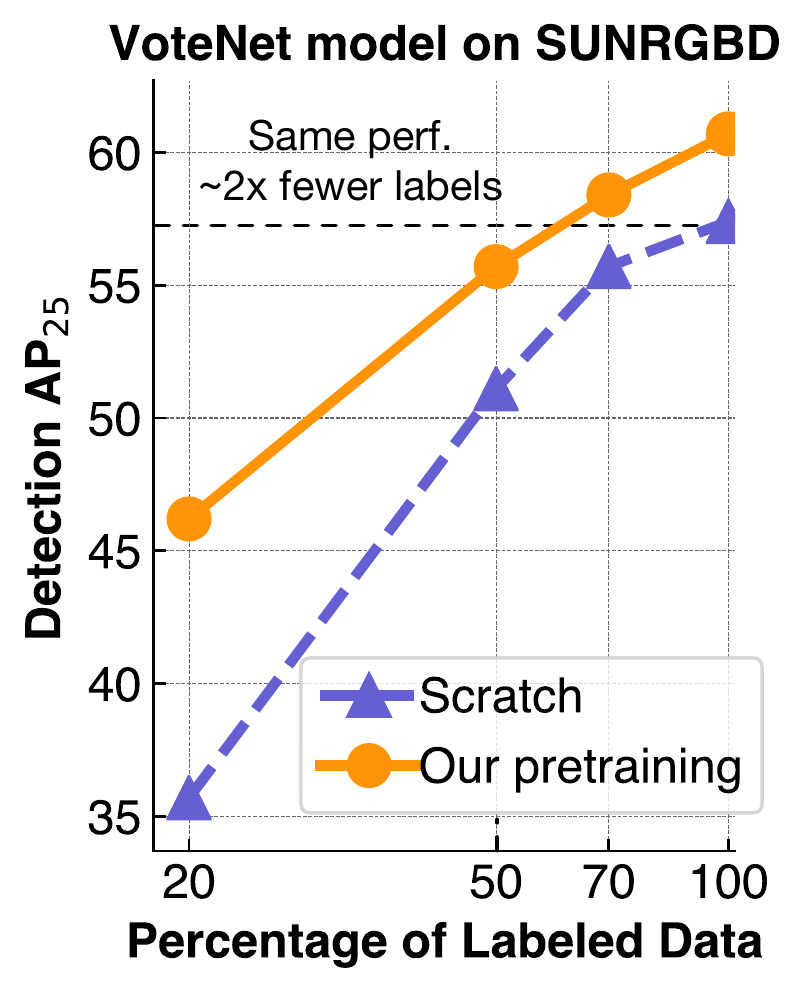}
	\vspace{-0.15in}
	\caption{\textbf{Label-efficiency of our self-supervised pretraining.}
		We finetune detection models from scratch or using our pretraining as initialization.
		Our pretraining which uses unlabeled single-view 3D data, outperforms training from scratch, and achieves the same detection performance with about half the detection labels.
	}
	\label{fig:teaser}
\end{figure}

Pretraining on depth maps faces several challenges: first, the absence of any supervision at scale requires the use of a self-supervised signal.
Second, the absence of alignments between depth maps excludes methods using multi-view constraints, such as finding correspondences between views~\cite{xie2020pointcontrast}.
Finally, the representation of a 3D scene varies depending on the application---segmentation typically uses a voxel representation~\cite{choy20194d}, whereas detection uses point-clouds~\cite{qi2019votenet}---and therefore the ideal pretraining method should be easily applicable to any 3D representation.

In this paper, we introduce a simple contrastive framework, {\bf \algorithmName}, to simultaneously learn different representations of any 3D data.
Our approach is based on the Instance Discrimination method by Wu \etal~\cite{wu2018unsupervised}, applying to depth maps.
We side-step the need of registered point clouds or correspondences, by considering each depth map as an instance and discriminating between them, even if they come from the same scene. For 3D representations with different architectures, we jointly learn features by considering the different representations as data augmentations that are processed with their associated networks~\cite{tian2019contrastive}.

Our contributions can be summarized as follows:
\begin{itemize}[leftmargin=*,itemsep=0em]
		\item We show that single view 3D depth scans can be used to learn powerful feature representations using self-supervised learning.
	\item We show that \emph{joint} training of different input representations like points and voxels is important for learning good representations, and a naive application of contrastive learning may not yield good results.
	\item Our method is applicable across different model architectures, indoor/outdoor 3D data, single/multi-view 3D data. We also show that it can be used to pretrain high capacity 3D architectures which otherwise overfit on tasks like detection and segmentation.
		\item We show performance improvements over \emph{nine} downstream tasks, and set a new state-of-the-art for \emph{two} object detection tasks (\scannet and \sunrgbd). Our models are efficient few-shot learners.
\end{itemize}

\begin{figure*}[t]
\centering
        \includegraphics[width=\linewidth]{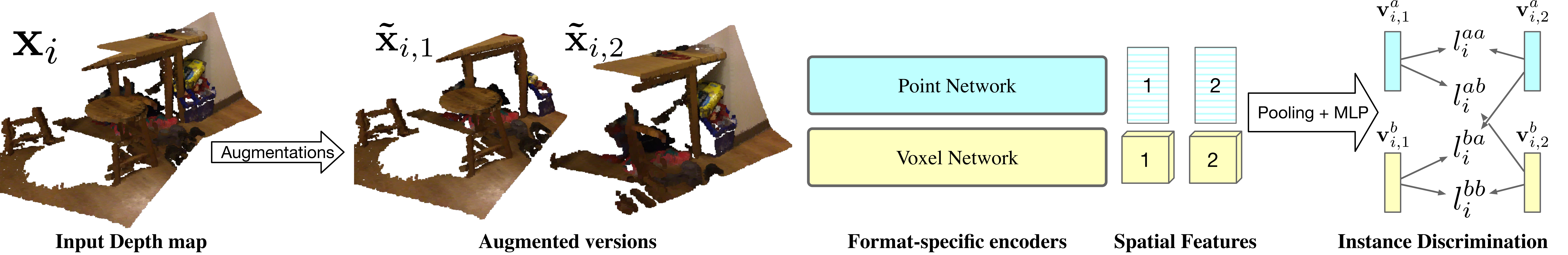}
    \vspace{-0.2in}
    \caption{\textbf{Approach Overview.} We propose \algorithmName{} - a simple 3D representation learning method that uses large amounts of unprocessed single/multi-view depth maps.
    Given a depth map we construct two augmented versions using data augmentation and represent them with different input formats (point coordinates and voxels).
    We use format-specific encoders to get spatial features which are pooled and projected to obtain global features $\bv$. The global features are used to setup an instance discrimination task and pretrain the encoders.}
    \label{fig:overview}
\end{figure*}

\vspace{-0.1in}
\section{Related Work}
\vspace{-0.05in}

Our method builds on the work from the self-supervised learning literature, with 3D data as an application. In this section, we give an overview of the recent advances in both self-supervision and 3D representations.
\vspace{0.02in}
\par \noindent \textbf{Self-supervised learning for images.}
Self-supervised learning is a well studied problem in machine learning and computer vision~\cite{vincent2008extracting,olshausen1996,ranzato2007unsupervised,salakhutdinov2009deep,masci2011stacked}.
There are many classes of methods for learning representations - clustering~\cite{bojanowski2017unsupervised,caron2018deep,ji2019invariant}, GANs~\cite{donahue2017adversarial,mescheder2017unifying}, pretext tasks~\cite{doersch2015unsupervised,noroozi2016unsupervised,wang2015unsupervised} \etc.
Recent advances~\cite{misra2019self,he2019momentum,henaff2019data,caron2020swav,chen2020simple,tian2020makes,kolesnikov2019revisiting,goyal2019scaling} have shown that self-supervised pretraining is a viable alternative to supervised pretraining for 2D recognition tasks.
Our work builds upon contrastive learning~\cite{hadsell2006dimensionality,oord2018representation} where models are trained to discriminate between each instance~\cite{dosovitskiy2016discriminative} with no explicit classifier~\cite{wu2018unsupervised}.
These instance discrimination methods can be extended to multiple modalities~\cite{tian2019contrastive,morgado2020avid,patrick2020multi}.
Our method extends the work of Wu \etal~\cite{wu2018unsupervised} to multiple 3D input formats following Tian et al.~\cite{tian2019contrastive} using a momentum encoder~\cite{he2019momentum} instead of a memory bank.
\vspace{0.02in}
\par \noindent \textbf{Self-supervised learning for 3D data.}
Most methods on self-supervised learning focus on single 3D object representation with different applications to reconstruction, classification or part segmentation~\cite{achlioptas2018learning,gadelha2018multiresolution,groueix2018papier,hassani2019unsupervised,li2018so,sauder2019self,wang2019deep,yang2018foldingnet,jing2020self,achituve2020self}.
Recently, Xie~\etal~\cite{xie2020pointcontrast} proposed a self-supervised method to build representations of scene level point clouds.
Their method relies on the complete 3D reconstruction of a scene with point-wise correspondences between the different views of a point cloud.
These point-wise correspondences requires post-processing the data by registering the different depth maps into a single 3D scene.
Their method can only be applied to static scenes that have been registered, which greatly limits the applications of their work.
We show a simple self-supervised method that learns state-of-the-art representations from \emph{single-view} 3D data, and can also be applied to multi-view data.
\vspace{0.01in}
\par \noindent \textbf{Representations of 3D scenes.}
There are multiple ways to represent 3D information in different vectorized forms such as point-clouds, voxels or meshes.
Point-cloud based models~\cite{qi2017pointnet,qi2017pointnet++} are widely used in classification and segmentation tasks~\cite{qi2017pointnet,qi2017pointnet++,klokov2017escape,xu2018spidercnn,wang2018dynamic,xie2018attentional,hua2018pointwise,atzmon2018point,tatarchenko2018tangent,wang2019associatively}, 3D reconstruction~\cite{tatarchenko2017octree,fan2017point,yang2018foldingnet} and 3D object detection~\cite{qi2019votenet,yi2019gspn,shi2019pointrcnn,yi2019gspn,pham2019jsis3d,wang2019associatively,zhang2020h3dnet}.
Since many 3D sensors acquire data in terms of 3D points, point clouds are a convenient input for deep networks.
However, since using convolution operations on point-clouds directly is difficult~\cite{qi2017pointnet,graham2015sparse}, voxelized data is another popular input representation.
3D convolutional models~\cite{tchapmi2017segcloud,riegler2017octnet,girdhar16learning,su2018splatnet,graham2015sparse,adams2010fast,hermosilla2018monte,li2018pointcnn,choy20194d} are widely used in 3D scene understanding~\cite{song2016deep,zhou2018voxelnet,graham20183d,yan2018second}.
There are also efforts to combine different 3D input representations~\cite{zhang2019path,zhang2020h3dnet,zhang2020deep,shi2020pv,gkioxari2019mesh,wang2018pixel2mesh}.
In this work, we propose to jointly pretrain two architectures for points and voxels, that are \pointnetplus~\cite{qi2017pointnet++} for points and Sparse Convolution based U-Net~\cite{choy20194d} for voxels.
\vspace{0.02in}
\par \noindent \textbf{3D transfer tasks and datasets.}
We use shape classification, scene segmentation, and object detection as the recognition tasks for transfer learning.
Shape classification techniques~\cite{qi2017pointnet, qi2017pointnet++, maturana2015voxnet, qi2016volumetric, su2015multi,chang2015shapenet} are widely evaluated on the ModelNet~\cite{WuSKYZTX15} dataset, which we use. It contains synthetic 3D data and each sample contains exactly one object.
We also evaluate on complete 3D scenes using the more general 3D scene understanding task.
Scene-centric datasets can be broadly divided into indoor scens~\cite{Silberman_ECCV12_NYUv2, sturm2012benchmark, xiao2013sun3d, song2015sun, armeni20163d, savva2016pigraphs, scenenn-3dv16, Dai_2017_CVPR_scannet, Matterport3D, pham2019jsis3d}, and outdoor (self-driving focussed) scenes~\cite{RosCVPR16,Geiger2013IJRR,sun2020scalability}.
We use these datasets and evaluate the performance of our methods on the indoor detection~\cite{qi2019votenet,zhang2020h3dnet,chen2020hierarchical,engelmann20203d,qi2018frustum}, scene segmentation~\cite{choy20194d,qi2017pointnet++,tchapmi2017segcloud,wu2019pointconv,yan2020pointasnl}, and outdoor detection tasks~\cite{shi2019pointrcnn,yan2018second,shi2020pv,shi2019points,li2019gs3d,yang2019std,chen2019fast}.

\vspace{-0.05in}
\section{Approach}
\label{sec:approach}
\vspace{-0.05in}

We develop a scalable pretraining method, \algorithmName, for 3D representations that uses unprocessed single-view or multi-view depth maps without human annotations.
Our method, illustrated in~\cref{fig:overview}, is based on the instance discrimination framework of Wu~\etal~\cite{wu2018unsupervised} with a momentum encoder~\cite{he2019momentum}.
\algorithmName learns 3D representations across multiple 3D input formats like points and voxels, and across different 3D architectures by using an extension of contrastive learning to multiple data formats~\cite{tian2019contrastive}.

\subsection{Instance Discrimination}

Given a dataset $\calD = \{\bx\}_{i=1}^{N}$ containing $N$ samples $\bx$, we wish to learn a function $g(\bx)$ that produces useful representations $\bv = g(\bx)$ of the input sample.
Our method uses 3D data where $\bx$ can be represented by point coordinates or voxels\footnote{Points in a depth map are a set, but for simplicity we denote them as a matrix. Our method does not rely on any specific ordering of the points.}.
We apply a data augmentation $t$ sampled randomly from a large set of augmentations $\calT$, to obtain an augmented sample $\btx = t(\bx)$.
The augmented sample is input to a deep network $g$ that extracts unit-norm global features $\bv = g(\btx)$ by pooling over the 3D spatial coordinates.
We setup an instance discrimination problem where the features $\bv_{i,1}$ and $\bv_{i,2}$ obtained from two data augmented versions of sample $i$ must be similar to each other, and different from features $\bv_j$ obtained using $K$ other (negative) samples $j$ in the dataset.
We use a contrastive loss~\cite{hadsell2006dimensionality,oord2018representation,sohn2016improved} to achieve this goal:
{\small
\begin{equation}
l_i = -\log\frac{\exp(\bv_{i,1}^{\top} \bv_{i,2}/\tau)}{\exp(\bv_{i,1}^{\top} \bv_{i,2}/\tau) + \sum_{j\neq i}^{K}\exp(\bv_{i,1}^{\top} \bv_j / \tau)}, \label{eqn:loss_sample}
\end{equation}
}

where $\tau$ is the temperature that controls the smoothness of the softmax distribution.
This loss encourages features from different augmentations of the same scene to be similar, while being dissimilar to features of other scenes. Thus, it learns features that focus on discriminative regions of a scene that make it different from other scenes in the dataset.

\par \noindent \textbf{Minimal assumptions on input data.}
Our method, by design, makes minimal assumptions about the input $\bx$, \ie, it is an unprocessed single-view depth map.
It does not require careful sampling of overlapping multi-view 3D inputs~\cite{xie2020pointcontrast} or object centric depth maps~\cite{hassani2019unsupervised,jing2020self}.
These minimal assumptions enable us to learn from large scale single-view 3D depth maps in~\cref{sec:experiments}, indoor and outdoor 3D depth maps obtained from a variety of sensors without relying on 3D calibration in~\cref{sec:outdoor}.

\par \noindent \textbf{Momentum encoder.}
As using a large number of negatives is important for contrastive learning~\cite{wu2018unsupervised,he2019momentum,misra2019self,chen2020simple}, we use the method of He \etal~\cite{he2019momentum} where the features of the other augmentation $\bv_{i,2}$ and negative samples $\bv_j$ in~\cref{eqn:loss_sample} are obtained using a momentum encoder and a queue respectively.
This allows us to use a large number $K$ of negative samples without increasing the training batch size.

\subsection{Extension to Multiple 3D Input Formats}
Multiple input formats are commonly used to represent 3D data -  point clouds, voxels, meshes \etc and have their specific deep learning architectures and applications.
Our self-supervised method can be naturally extended to accommodate these input formats and architectures.
For each input format $f$, we denote the corresponding input sample as $\bx^{f}$, the format-specific encoder network as $g^{f}$, and the extracted feature as $\bv^{f}$.
Extending~\cref{eqn:loss_sample}, we can minimize a single objective that performs instance discrimination within and across input formats $a, b$:
{\small
\begin{equation}
l^{ab}_i = -\log\frac{\exp(\bv_{i,1}^{a\top} \bv^{b}_{i,2}/\tau)}{\exp(\bv_{i,1}^{a\top} \bv^{b}_{i,2}/\tau) + \sum_{j\neq i}^{K}\exp(\bv_{i,1}^{a\top} \bv^{b}_j / \tau)}. \label{eqn:loss_sample_formats}
\end{equation}
}

When the input formats $a, b$ are identical, this objective reduces to the \emph{within format} loss of~\cref{eqn:loss_sample}, and when $a\!\neq\!b$ this objective aligns the feature representations $\bv^{f} = g^{f}(\bx^{f})$ obtained \emph{across formats} $f$ using different network architectures $g^{f}$.
As illustrated in~\cref{fig:overview}, we use two popular input formats - point clouds and voxels, and train these format-specific models with a single joint loss function
{\small
\begin{equation}
L_{i} = \underbrace{l^{ab}_{i} + l^{ba}_{i}}_\text{across format} + \underbrace{l^{aa}_{i} + l^{bb}_{i}}_\text{within format}. \label{eqn:loss_final}
\end{equation}
}

Similar techniques have been explored in the context of different modalities of data, \eg, color and grayscale images~\cite{tian2019contrastive}, audio and video~\cite{morgado2020avid,patrick2020multi} \etc.
While these methods use different modalities, our extension uses the \emph{same 3D data} and only changes the input format.

\subsection{Model Architecture}
\label{sec:model_arch}
We describe the model architecture used for our input format-specific encoders.
Once we obtain the augmented data sample $\tilde{\bx}$, we use the XYZ coordinates for point input.
We follow~\cite{choy20194d,xie2020pointcontrast} to voxelize the augmented sample $\tilde{\bx}$ and use the 3D voxel occupancy grid and RGB values for the voxel model.
Thus, both encoders operate on the same input 3D data, and differ only in the way the input is represented.
We provide the full, layer-wise architecture details in the supplemental material.

\par \noindent \textbf{Point input.} We use \pointnetplus~\cite{qi2019votenet} as the backbone network which takes as input the XYZ coordinates of the 3D data.
\pointnetplus employs a U-Net structure which has four layers of feature extraction and down-sampling, and two layers of feature aggregation and up-sampling.
Our network takes as input $20$K points from the scene represented as a $20$K$\times 3$ matrix of XYZ coordinates.
The network's final layer produces $C$ dimensional per-point features for $1024$ points after aggregation.
We obtain the scene level $256$ dimensional feature $\bv$ in~\cref{eqn:loss_sample_formats} by global max pooling to these last layer features, followed by a two layer MLP as in~\cite{chen2020simple} and L2 normalization.
We increase the size $C$ of the output feature in the last two upsampling (\texttt{fp}) layers from $256$ in~\cite{qi2019votenet} to $512$.
This improves the performance of the baselines, \eg, training from scratch, as well as our method and we use this improved architecture throughout the paper.

\par \noindent \textbf{Voxel input.}
We use a sparse convolution U-Net model~\cite{choy20194d} as the backbone for the voxel 3D input.
The network takes a 3D occupancy grid as the input representation of the 3D data.
Our U-Net consists of four layers of feature extraction and pooling and four layers of feature aggregation and up-sampling.
We use a voxel size of $5cm$ to voxelize the input data and following~\cite{choy20194d}, use the voxel occupancy grid and the RGB values as input to the model.
To obtain the scene level $256$ dimensional feature $\bv$ (~\cref{eqn:loss_sample_formats}), similar to the point input, we apply global max-pooling to the last layer feature, followed by a two layer MLP and L2 normalization.

\par \noindent \textbf{Using global feature representations.}
Our instance discrimination formulation learns global feature representations $\bv$ from the input 3D data $\bx$, \ie, the feature representation $\bv$ is obtained by pooling over the 3D coordinates of the input.
On the other hand many downstream 3D tasks require local per-point predictions like segmentation.
We show in~\cref{sec:experiments} that despite using global features, our pretraining benefits such prediction tasks.

\subsection{Data Augmentation for 3D}
\label{sec:data_aug}
Data augmentation is as an essential component of our framework. We first adopt standard pointcloud data argumentation methods proposed in ~\cite{qi2019votenet}, which are random point up/down sampling, random flip in xy axis, and random rotation. However, after adding these methods, it is still easy for the network to distinguish different training instances. Thus, we add two new data augmentation methods: random cuboid and random drop patches. Inspired by the random crop in 2D images~\cite{szegedy2015going}, we define a random cuboid augmentation that extracts random cuboids from the input point cloud. Cuboids are sampled using a random scale $[0.5, 1.0]$ of the original scene, and a random aspect ratio $[0.75, 1.0]$. We also drop (erase) cuboids to force the network learn local geometric features. The dropped cuboid is randomly cropped with $0.2$ of the scene scale.
The performance boost from each augmentation is analyzed in~\cref{sec:analysis}.
For voxelized inputs, in addition to all the point augmentations, we use the augmentations from~\cite{choy20194d}.

\subsection{Implementation Details}
\label{sec:implementation_details}
We use $130$K negatives for contrastive learning in~\cref{eqn:loss_final} and a momentum of $0.9$ for the momentum encoder following~\cite{he2019momentum}.
As noted in~\cref{sec:model_arch}, we follow Chen \etal~\cite{chen2020simple} and use an additional non-linear projection (two layer MLP) and L2 normalization after the network $g$ to obtain the features $\bv$. The features $\bv$ are $128$ dimensional and we use a temperature value of $0.1$ while computing the non-parametric softmax in~\cref{eqn:loss_sample}. We use a standard SGD optimizer with momentum $0.9$, cosine learning rate scheduler~\cite{loshchilov2016sgdr} starting from $0.12$ to $0.00012$ and train the model for $1000$ epochs with a batch size of $1024$.

\section{Experiments}
\label{sec:experiments}

\begin{table}
\centering
\setlength{\tabcolsep}{0.1em}\resizebox{\linewidth}{!}{
\begin{threeparttable}
\begin{tabular}{l|ccr}
\toprule
\textbf{Dataset} & \textbf{Stats} & \textbf{Task} & \textbf{Gain of}\\
& & & \textbf{\algorithmName} \\
\shline
\rowcolor{highlightRowColor} \multicolumn{4}{c}{Self-supervised Pretraining} \\
\shline

\scanVideo~\cite{Dai_2017_CVPR_scannet} & \multicolumn{3}{l}{~~190K single-view depth maps (Indoor)} \\

\hline
\redVideo~\cite{Choi2016} & \multicolumn{3}{l}{~~370K single-view depth maps (Indoor/Outdoor)} \\

\shline
\rowcolor{highlightRowColor} \multicolumn{4}{c}{Transfer tasks} \\
\shline

\scannet~\cite{Dai_2017_CVPR_scannet} & 1.2K train, 312 val (Indoor) & Det. & \textbf{+3.6\% mAP}\\
&  & Seg. & \small{\textbf{+0.9\% mIOU}$\dagger$}\\

\hline
\sunrgbd~\cite{song2015sun} & 5.2K train, 5K val (Indoor) & Det & \small{\textbf{+3.3\% mAP}} \\

\hline

\stanford~\cite{2D-3D-S} & 199 train, 67 val (Indoor) & Det & \small{\textbf{+12.1\% mAP}} \\
& & Seg. & \small{\textbf{+2.4\% mIOU}}\\

\hline
\synthia~\cite{RosCVPR16} & 19.8K train, 1.8K val (Synth.) & Seg. & \small{\textbf{+2.4\% mIOU}}\\

\hline
\matterport~\cite{Matterport3D} & 1.4K train, 232 val (Indoor) & Det. &  \small{\textbf{+3.9\% mAP}}\\

\hline
\modelnet~\cite{WuSKYZTX15} & 9.8K train, 2.4K val (Synth.) & Cls. & \small{\textbf{+3.1\% Acc}$^\dagger$}\\
\bottomrule
\end{tabular}
\begin{tablenotes}[flushleft]
\item[] Det.: Object Detection, Seg: Semantic Segmentation
\item[] Cls: Classification, Synth.: Synthetic, $^\dagger$Results in supplemental.
\end{tablenotes}
\end{threeparttable}
}
\vspace{-0.15in}
\caption{Pretraining datasets and transfer tasks used in this paper. 
We use two different pretraining datasets without post-processing like 3D registration, camera calibration.
We use $8$ different transfer tasks for evaluation where our \algorithmName pretraining gives consistent gains (last column) over scratch pretraining. Additonally, we show evaluation results on LiDAR data in~\cref{sec:outdoor}.
}
\label{tab:datasets_and_tasks}
\end{table}

We evaluate \algorithmName pretraining by transfer learning, \ie, fine-tuning on downstream tasks and datasets.
As ~\cref{tab:datasets_and_tasks} shows, we use a diverse set of 3D understanding tasks like object classification, semantic segmentation, and object detection.
We first study a single input 3D format and a single network architecture in~\cref{sec:point_vs_prior_work} showing that \algorithmName's performance improves with large data and higher capacity models, and benefits finetuning with limited labels.
Finally, in~\cref{sec:multiple_input_formats}, we show the benefits of our pre-training across different 3D input formats (points and voxels).

\vspace{0.05in}
\par \noindent \textbf{Pretraining Details.}
We use single-view depth map videos from the popular \scannet\cite{Dai_2017_CVPR_scannet} dataset and term it as \scanVideo.
\scanVideo contains about 2.5 million RGB-D scans for more than 1500 indoor scenes.
Following the train/val split from~\cite{qi2019votenet}, we extract around 190K RGB-D scans (one frame every 15 frames) from about 1200 video sequences in the train set.
We do not use camera calibration or 3D registration methods and operate directly on single-view depth maps.
We use our data augmentation described in~\cref{sec:data_aug} and use the training objectives from~\cref{sec:approach}.
Additional details are provided in~\cref{sec:implementation_details} and the supplemental material.

\begin{table}[!t]
	\centering
\setlength{\tabcolsep}{0.2em}\resizebox{\linewidth}{!}{
	\begin{tabular}{@{}lccccccccccc@{}}
	\toprule
		 \textbf{Initialization} & \textbf{\scannet} & \textbf{\sunrgbd} & \textbf{\matterport} & \textbf{\stanford} \\
		\hline
	Scratch & 58.6 & 57.4 & 38.8 & 31.2\\
	Supervised & - & \phantom{\tabgreen{(+1.7)}}59.1 \tabgreen{(+1.7)} & \phantom{\tabgreen{(+1.7)}}41.7 \tabgreen{(+2.9)} & \phantom{\tabgreen{(+17.3)}}\textbf{48.5} \tabgreen{(+17.3)} \\
		\algorithmName (Ours) & \phantom{\tabgreen{(+1.7)}}\textbf{61.3} \tabgreen{(+2.7)} & \phantom{\tabgreen{(+1.7)}}\textbf{60.4} \tabgreen{(+3.0)} & \phantom{\tabgreen{(+1.7)}}\textbf{41.9} \tabgreen{(+3.1)} & \phantom{\tabgreen{(+12.1)}}43.3 \tabgreen{(+12.1)} \\
	\hline
	PointContrast~\cite{xie2020pointcontrast} & 59.2 & 57.5 & - & - \\
		\bottomrule
	\end{tabular}
}
\vspace{-0.1in}
\caption{\textbf{Detection AP$_{25}$ using VoteNet~\cite{qi2019votenet}.} We evaluate different pretrained models - random initialization, supervised VoteNet on \scannet, and our self-supervised \algorithmName using the point input format.
\algorithmName outperforms the scratch model on all benchmarks and is better than the detection-specific supervised pretraining on two datasets.
}
\label{tab:pointnet_scannet}
\end{table}

\begin{table}[!t]
	\centering
\setlength{\tabcolsep}{0.4em}\resizebox{\linewidth}{!}{
	\begin{tabular}{@{}lccccccccccc@{}}
	\toprule
			\textbf{Method} & \multicolumn{2}{c}{\textbf{\scannet}} & \multicolumn{2}{c}{\textbf{\sunrgbd}}\\
	& \textbf{AP}$_{25}$ & \textbf{AP}$_{50}$ & \textbf{AP}$_{25}$ & \textbf{AP}$_{50}$ \\
		\hline
	F-PointNet~\cite{qi2018frustum} & 54.0 & - & - & -\\
	VoteNet~\cite{qi2019votenet} & 58.6 & 33.5 & 57.7 & 32.9\\
	H3DNet~\cite{zhang2020h3dnet} & 67.2 & 48.1 & 60.1 & 39.0 \\
	HGNet~\cite{chen2020hierarchical} & 61.3 & 34.4 & 61.6 & 34.4 \\
	3D-MPA~\cite{engelmann20203d} & 64.2 & 49.2 & - & - \\
	PointContrast (VoteNet)~\cite{xie2020pointcontrast} & 59.2 & 38.0 & 57.5 & 34.8 \\
	\hline
	\algorithmName (VoteNet) & 64.0 & 42.9 & 61.6 & 35.5 \\
	\algorithmName (H3DNet) & \textbf{69.0} & \textbf{50.0} & \textbf{63.5} & \textbf{43.4}\\
	\bottomrule
	\end{tabular}
}
\vspace{-0.1in}
\caption{\textbf{Transfer using state-of-the-art detection frameworks.} We use our pretrained model (\pointnetplus $3\times$ on \redVideo+\scanVideo) and transfer it using two state-of-the-art detection frameworks - H3DNet~\cite{zhang2020h3dnet} and VoteNet~\cite{qi2019votenet}.
Our \algorithmName pretraining outperforms all prior work and sets a new state-of-the-art on both \scannet and \sunrgbd detection datasets.
}
\label{tab:detection_best}
\end{table}

\begin{figure*}
\centering
\includegraphics[width=0.95\textwidth]{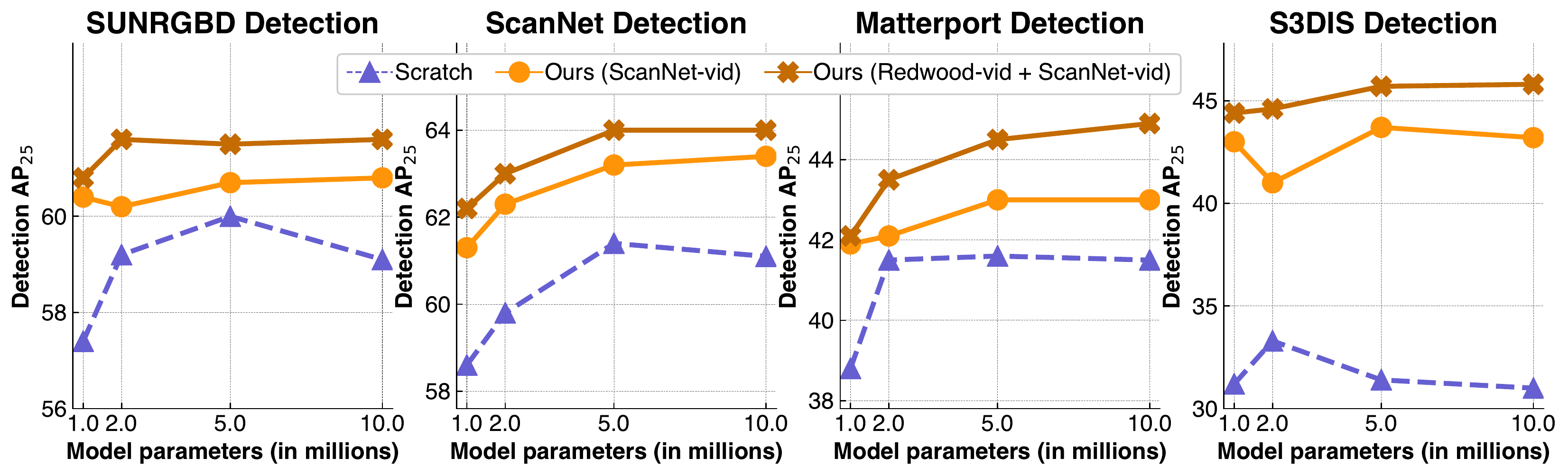}
\vspace{-0.1in}
\caption{\textbf{Scaling the model size and pretraining data.} We increase the model capacity of the \pointnetplus model by increasing the width by $\{2\times,3\times,4\times\}$.
When training from scratch, increasing the model capacity increases the performance but ultimately leads to overfitting. Overfitting is more pronounced on small datasets like \stanford. Our \algorithmName pretraining on \scanVideo improves the performance for larger models and reduces overfitting. We increase the pretraining data by combining the readily available single-view depth maps from \scanVideo and \redVideo. \algorithmName's performance improves significantly when using both large data and large models.
}
\label{fig:large_point_models}
\end{figure*}

\vspace{0.05in}
\subsection{Transfer Datasets and Tasks}
We evaluate our pretrained model by transfer learning and finetune it on different downstream datasets and tasks summarized in~\cref{tab:datasets_and_tasks}.
We use diverse downstream datasets - full scenes/object centric; using different 3D sensors; single/multi-view; real/synthetic; indoor/outdoor.
On these diverse datasets, we use three major tasks, which are classification, semantic segmentation, and object detection.
These tasks test different aspects of the pretrained model - while object detection and semantic segmentation use local features, classification is performed on global features.

\subsection{Pretraining with Point Input Format}
\label{sec:point_vs_prior_work}

We pretrain a \pointnetplus model using the instance discrimination objective in~\cref{eqn:loss_sample} on the single-view depth maps from \scanVideo.
We study the transfer performance of the pretrained model on object detection using the VoteNet~\cite{qi2019votenet} framework that uses a \pointnetplus backbone network.
In~\cref{tab:pointnet_scannet} we report the detection results by finetuning the VoteNet model with different backbone initializations.
We use the implementation of~\cite{qi2019votenet} for finetuning and report the detection performance using the mean Average Precision at IoU=0.25 (AP$_{25}$) metric.
 Training from scratch or random initialization is standard practice in VoteNet~\cite{qi2019votenet} and serves as a baseline for comparing other pretraining methods.
As a supervised pretraining baseline, we use the VoteNet model trained on \scannet detection.
Since the supervised baseline is pretrained specifically on object detection, it serves as a strong baseline.

Scratch training provides competitive results on the larger detection datasets like \scannet and \sunrgbd~\cite{song2015sun,silberman2012indoor,janoch2013category,xiao2013sun3d}, however, its performance on the smaller \stanford dataset is low.
In comparison, supervised pretraining provides large gains in the detection performance across all datasets.
\algorithmName outperforms training from scratch on all the four datasets, and \textbf{improves performance by 12.1\% mAP} on the small \stanford dataset that has only 200 labeled training samples.
We further analyze label efficiency of our model in~\cref{sec:label_efficiency}.
Interestingly, despite using no labels during pretraining, \algorithmName is better than the detection-specific supervised pretraining for two datasets (\sunrgbd and \matterport).
Our method also outperforms recent work~\cite{xie2020pointcontrast} by significant margins.

\vspace{-0.15in}
\subsubsection{Training Higher Capacity Models}
\label{sec:larger_models}

We use \algorithmName for training higher capacity models.
We follow the practice in 2D self-supervised learning~\cite{kolesnikov2019revisiting} and increase the capacity of the \pointnetplus model by multiplying the channel width of all the layers by $\{2, 3, 4\}$.
We pretrain all models on the \scanVideo dataset and measure their transfer performance in~\cref{fig:large_point_models}.
Training large models from scratch provides some benefit, but quickly leads to reduced or plateauing performance.
We observe overfitting on the small datasets like \stanford where increasing the model capacity does not improve performance.
On the other hand, our self-supervised pretraining on \scanVideo reduces this overfitting and performance improves or stays the same for larger models.
This suggests that \emph{pretraining is crucial} for training large 3D detection models.

\vspace{-0.1in}
\subsubsection{Using More Pretraining Data}
\label{sec:larger_data}
\vspace{-0.05in}
We increase the pretraining data by using readily available single-view 3D data from the \redVideo dataset~\cite{Choi2016}.
\redVideo contains over 23 million depth scans from RGB-D videos taken in both indoor and outdoor settings.
As this dataset is extremely large, we use a subset of 2500 video sequences consisting of 10 categories and extract 370K RGB-D scans.
More importantly, the \redVideo dataset does \emph{not contain} camera extrinsic parameters and thus cannot be registered to get a multi-view dataset which is a necessity for prior self-supervised methods~\cite{xie2020pointcontrast}.

Combining the \redVideo and \scanVideo datasets allows us to triple our pretraining data.
We pretrain all models on this combined dataset and report their performance (AP$_{25}$) in~\cref{fig:large_point_models}.
\algorithmName's performance improves with both model capacity and number of pretraining samples across all four detection datasets.
The higher capacity models show a larger improvement in performance particularly on the smaller \stanford dataset.
This suggests that \algorithmName can leverage large amounts of single-view 3D data to obtain better and higher capacity 3D models.

\vspace{-0.1in}
\subsubsection{State-of-the-art Detection Frameworks}
\vspace{-0.05in}
We use two state-of-the-art detection frameworks - H3DNet~\cite{zhang2020h3dnet} and VoteNet~\cite{qi2019votenet} and study the benefit of using our pretrained model.
We use our \pointnetplus $3\times$ model pretrained on the combined \redVideo and \scanVideo dataset and transfer it using these detection frameworks.
The detection results in~\cref{tab:detection_best} show that our pretrained model achieves state-of-the-art performance on \sunrgbd and \scannet.
In particular, as the gains are larger on stricter mAP at IoU=0.5, our pretrained models result in detection models that are better at localization.

\begin{figure}[!t]
	\centering
	\includegraphics[width=\columnwidth]{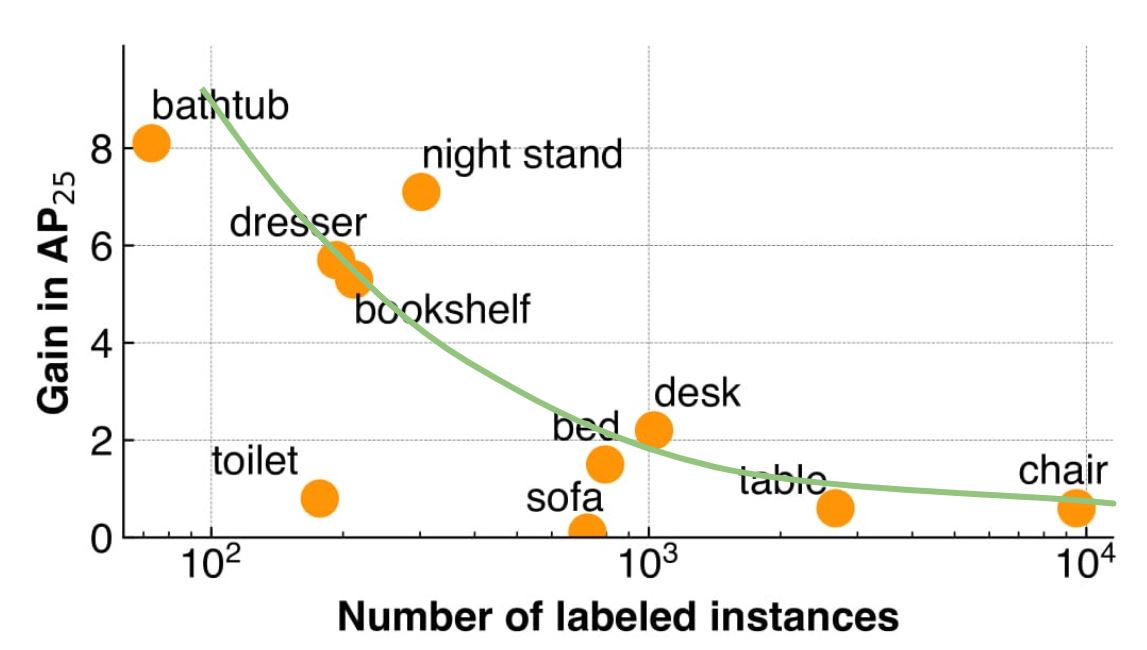}
	\vspace{-0.2in}
	\caption{\textbf{Pretraining benefits long tail classes.} We analyze the gain of our pretraining across different classes for \sunrgbd object detection.
	The training data has a long tailed distribution where the least frequent classes occur $50\times$ less than the most frequent classes.
	Our pretraining improves performance for classes with fewer labeled instances by $4-8\%$. (Trending line in green.)}
	\label{fig:class_vs_perform}
\end{figure}

\vspace{-0.1in}
\subsubsection{Label Efficiency of Pretrained Models}
\label{sec:label_efficiency}
\vspace{-0.05in}

Pretraining allows models to be finetuned with small amount of labeled data.
In~\cref{tab:pointnet_scannet}, we observe that small labeled datasets benefit more from pretraining.
We study the label efficiency of \algorithmName pretrained models by varying the amount of labeled data used for finetuning.
While varying the data, we draw 3 independent samples and report average results.
We use the \pointnetplus models pretrained on \scanVideo (~\cref{sec:point_vs_prior_work}) and report the detection performance in~\cref{fig:teaser}.
\algorithmName pretraining provides large gains in performance at every setting.
On both the \scannet and \sunrgbd datasets, our model with just \emph{$50\%$ samples gets the same performance} as training from scratch with the full dataset.
When using $20\%$ samples for finetuning, our pretrained models provide a \textbf{gain of over 10\% mAP}.
This shows that our pretraining is label efficient and can improve performance especially on tasks with limited supervision.
\par \noindent \textbf{Does pretraining benefit tail classes?} 3D detection datasets like \sunrgbd and \scannet exhibit a long tailed distribution where many `tail' classes have few training instances.
In the \sunrgbd dataset, the `tail' classes like \texttt{bathtub}, \texttt{toilet}, \texttt{dresser} have less than 200 training instances, while classes like \texttt{chair} have over 9000 instances.
\cref{fig:class_vs_perform} shows the gain of our pretrained model over the scratch model across object classes on the \sunrgbd dataset.
Our pretraining improves the performance of classes with fewer instances, \ie, the tail classes, by $4-8\%$ AP.
This suggests that \algorithmName pretraining can partially address the long tailed label distributions of current 3D scene understanding benchmarks.

\vspace{-0.05in}
\subsection{Pretraining with Multiple Input Formats}
\label{sec:multiple_input_formats}
\vspace{-0.05in}

\begin{table}[!t]
	\centering
	\setlength{\tabcolsep}{0.2em}\resizebox{\linewidth}{!}{
		\begin{tabular}{@{}l|cc|cc@{}}
			\toprule 
			\textbf{Loss} & \multicolumn{2}{c}{\textbf{Point Transfer}} & \multicolumn{2}{c}{\textbf{Voxel Transfer}} \\
			& \textbf{\sunrgbd} & \textbf{\scannet} & \textbf{\stanford} & \textbf{\synthia} \\
			\hline
			Scratch & 57.4 & 58.6 & 68.2 & 78.9 \\
			\hline
			Within Format only & \phantom{\tabgreen{(+3.0)}}60.4 \tabgreen{(+3.0)} & \phantom{\tabgreen{(+1.7)}} 61.3 \tabgreen{(+1.7)}& \phantom{\tabred{(-2.7)}}66.5 \tabred{(-2.7)}& \phantom{\tabgreen{(+1.2)}} 80.1 \tabgreen{(+1.2)}\\
			Across format only & \phantom{\tabgreen{(+2.6)}} 60.0 \tabgreen{(+2.6)} & \phantom{\tabgreen{(+2.5)}} 61.1 \tabgreen{(+2.5)} & \phantom{\tabgreen{(+1.7)}} 69.9 \tabgreen{(+1.7)} & \phantom{\tabgreen{(+2.3)}} 81.2 \tabgreen{(+2.3)}\\
			Both (Ours) & \phantom{\tabgreen{(+3.3)}} \textbf{60.7} \tabgreen{(+3.3)} & \phantom{\tabgreen{(+3.6)}} \textbf{62.2} \tabgreen{(+3.6)} & \phantom{\tabgreen{(+2.4)}} \textbf{70.6} \tabgreen{(+2.4)} & \phantom{\tabgreen{(+2.4)}} \textbf{81.3} \tabgreen{(+2.4)} \\
			\hline
			PointContrast~\cite{xie2020pointcontrast} & 57.5 & 59.2 & 70.9 & 83.1 \\
			\bottomrule
		\end{tabular}
	}
	\vspace{-0.1in}
	\caption{\textbf{Multiple input formats.} We study the importance of training 3D representations jointly using multiple input formats - points and voxels.
	We vary the within format and across format loss terms in~\cref{eqn:loss_final}.
	We report detection mAP@0.25 on the point transfer tasks and segmentation mIOU for the voxel transfer tasks.
	A naive within format instance discrimination loss (second row) can give worse performance than finetuning from scratch.
	We observe that performing instance discrimination across the input formats (third row) greatly improves over the within format loss term.
	Our joint loss gives the best performance.}
	\label{tab:analysis_formats}
\end{table}
We pretrain \algorithmName using both the point and voxel input formats and use two format-specific encoders - \pointnetplus for points and \mink for voxels.
As explained in~\cref{eqn:loss_final}, when using multiple 3D input formats, we can define two loss terms - a within format loss and an across format loss.
To analyze which of these loss terms matter for pretraining, we consider three variants - \emph{(1) Within format} which independently trains format-specific models for each input format and is a straightforward application of instance discrimination to 3D;
\emph{(2) Across format} which trains the format-specific models jointly using the second term of~\cref{eqn:loss_final};
\emph{(3) Ours} which trains the format-specific models jointly using our combined loss function.
We evaluate the pretrained models by transfer learning.
As in~\cref{sec:point_vs_prior_work}, we finetune the pretrained point input format \pointnetplus models on \sunrgbd and \scannet detection using the VoteNet framework.
We finetune the voxel \mink models on segmentation using the framework from Spatio-Temporal Segmentation~\cite{choy20194d} which uses a \mink backbone network. The results are summarized in~\cref{tab:analysis_formats}.

Compared to training from scratch, the within format pretraining only provides a benefit for the point input format \pointnetplus models.
For the voxel models, this pretraining does not improve consistently over training from scratch, which is in line with observations from recent work~\cite{xie2020pointcontrast}.
This shows that a naive application of instance discrimination to 3D representation learning \emph{may not} yield good pretrained models.
The across format loss improves performance for both the point and voxel models, suggesting the benefit of using multiple input formats.
Our proposed loss function that combines both the within and across format losses provides the best transfer performance.
The gains are particularly significant on the voxel format model which improves by \textbf{$4\%$ over the within format loss}.
In the supplemental material, we show that this benefit of joint training over the within format loss also holds across different pretraining data and architectures.
Although our method only uses single-view unprocessed depth scans, our results on the voxel transfer tasks are comparable to the recent PointContrast method~\cite{xie2020pointcontrast} that uses multi-view point clouds and pointwise correspondences.
We note that our \mink architecture is different from~\cite{xie2020pointcontrast} since their architecture underfit on our self-supervised pretraining task.
We provide results with their architecture in the supplement.

\section{Analysis}
\label{sec:analysis}

In this section we present a series of experiments designed to understand \algorithmName better.
We first pretrain point format (\pointnetplus) models on the \scanVideo dataset following the settings from~\cref{sec:point_vs_prior_work}.
We use two transfer tasks for evaluation - (1) object detection on \sunrgbd using VoteNet~\cite{qi2019votenet} where we finetune the full model and test the quality of the pretraining; (2) object classification on \modelnet where we keep the model fixed and only train linear classifiers on fixed features, thus testing the quality of the learned representations~\cite{goyal2019scaling,kolesnikov2019revisiting}. Finally, we also evaluate \algorithmName's generalizability to outdoor 3D data.

\begin{table}[!t]
	\centering
	\setlength{\tabcolsep}{0.2em}\resizebox{\linewidth}{!}{
																		
		\begin{tabular}{@{}l|ccc@{}}
			\toprule
			\textbf{Task} & VoteNet~\cite{qi2019votenet} & +Rand. & +Rand. \\
			 & & Cuboid & Drop \\
			\hline
			\textbf{\modelnet Linear} (Accuracy)& 80.6 & \textbf{85.4} & \uline{85.0} \\
			\textbf{\sunrgbd Detection} (mAP)& 58.6 & 59.5 & \textbf{60.7}\\
			\bottomrule
		\end{tabular}
	}
	\vspace{-0.1in}
	\caption{\textbf{Data augmentation.} We vary the data augmentation used for pretraining \algorithmName point models and report their transfer performance. The standard data augmentation used in supervised learning (VoteNet) is not sufficient to learn good self-supervised models. Our proposed Random Cuboid and Random Drop augmentations improve performance.}
	\label{tab:analysis_data}
\end{table}

\subsection{Importance of Data Augmentation}
\label{sec:analyze_data_aug}
Data augmentations play an important role for self-supervised representation learning and have been studied extensively in the case of 2D images~\cite{caron2020swav,misra2019self,chen2020improved,tian2020makes,tian2019contrastive}.
However, the impact of data augmentation for 3D representation learning is less well understood.
Thus, we analyze the effect of our proposed augmentations from~\cref{sec:data_aug} on transfer performance.
We train different \algorithmName point models with the same training setup and only vary the data augmentation used. Our results are summarized in~\cref{tab:analysis_data}.

The widely used VoteNet~\cite{qi2019votenet} augmentations from supervised learning perform worse than our proposed augmentations.
Our augmentations lead to both a better feature representation: a gain of $5\%$ accuracy on \modelnet classification, and a better pretrained model: $2\%$ mAP on \sunrgbd detection. In our experiments, we consistently observe gains from our improved data augmentation on all the downstream tasks from~\cref{sec:experiments} which underscores the importance of designing good data augmentation.

\subsection{Impact of Single-view or Multi-view 3D Data}
\label{sec:analyze_single_view}

\begin{table}[!t]
	\centering
	\setlength{\tabcolsep}{0.2em}\resizebox{\linewidth}{!}{
		\begin{tabular}{@{}l|c|ccc@{}}
			\toprule
			& \multicolumn{4}{c}{\textbf{Pretraining}} \\
			\textbf{Task} & Scratch & \scannet & \scanVideo & \redVideo\\
			& & \small{(Multi-view)} & \small{(Single-view)} & \small{(Single-view)} \\
			\hline
			\textbf{\modelnet Linear} (Accuracy)& 50.7 & 85.1 & 85.0 & 86.4 \\
			\textbf{\sunrgbd Detection} (mAP)&  57.4 & 60.5 & 60.7 & 60.4\\
			\bottomrule
		\end{tabular}
	}
	\vspace{-0.1in}
	\caption{\textbf{Single-view or multi-view 3D data.} We study whether our pretraining is sensitive to single-view or multi-view data.
	We use \scannet and \scanVideo which are multi-view and single-view versions of the same dataset~\cite{Dai_2017_CVPR_scannet}, and \redVideo~\cite{Choi2016} which is a single-view only 3D dataset. Our pretrained model is robust to 3D preprocessing, and using single-view or multi-view data gives similar performance.
	}
	\label{tab:analysis_aligned}
\end{table}
Our self-supervised method does not make assumptions on the input data and can use single-view depth maps without 3D preprocessing as input.
We study whether pretraining on reconstructed multi-view 3D scenes impacts the downstream performance.
We use the \scannet~\cite{Dai_2017_CVPR_scannet} dataset which contains multi-view 3D data obtained by 3D registration of the \scanVideo depth maps.
As another single-view dataset, we pretrain on the \redVideo dataset from~\cref{sec:larger_data}.
We pretrain \algorithmName point models on these datasets and compare their performance by transfer learning in~\cref{tab:analysis_aligned}.

The transfer performance is similar when models are pretrained on \scanVideo or \scannet.
Since \scanVideo and \scannet only differ in the 3D preprocessing involved, the result suggests that \algorithmName is not sensitive to single-view or multi-view input data.
This is not surprising given that our objective does not rely on multi-view information.
Pretraining on the single-view \redVideo dataset also gives similar performance suggesting that \algorithmName is robust to different data distributions during pretraining.
All the \algorithmName models outperform the scratch model.

\begin{figure}
\centering
\includegraphics[width=0.23\textwidth]{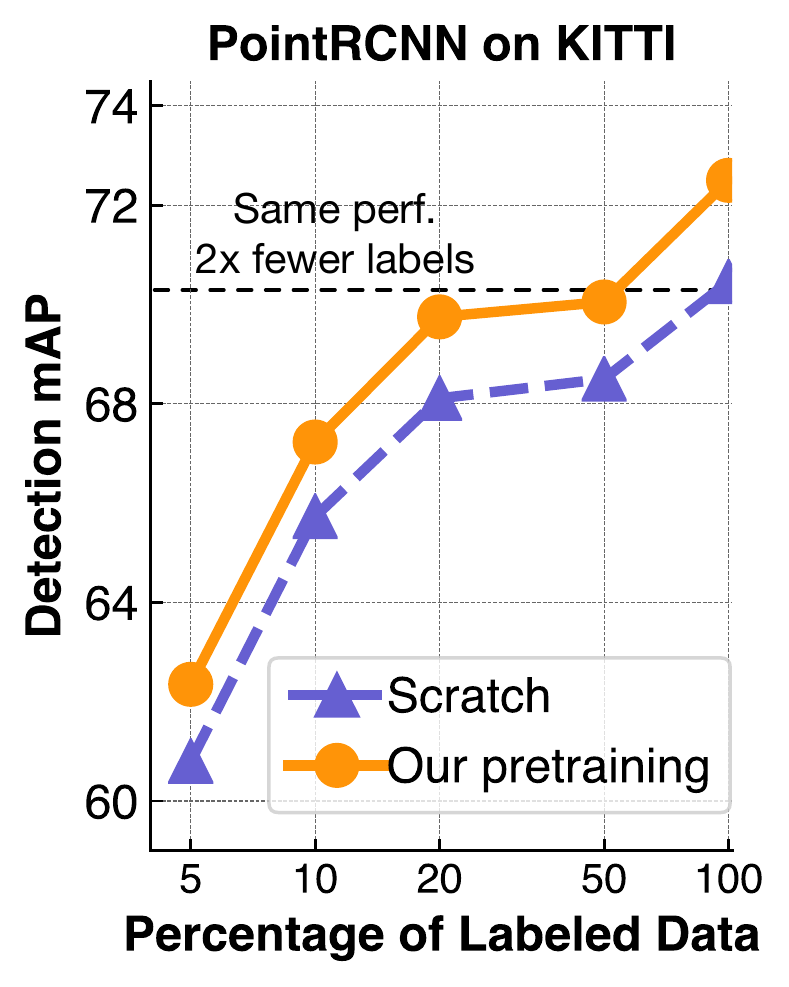}
\includegraphics[width=0.23\textwidth]{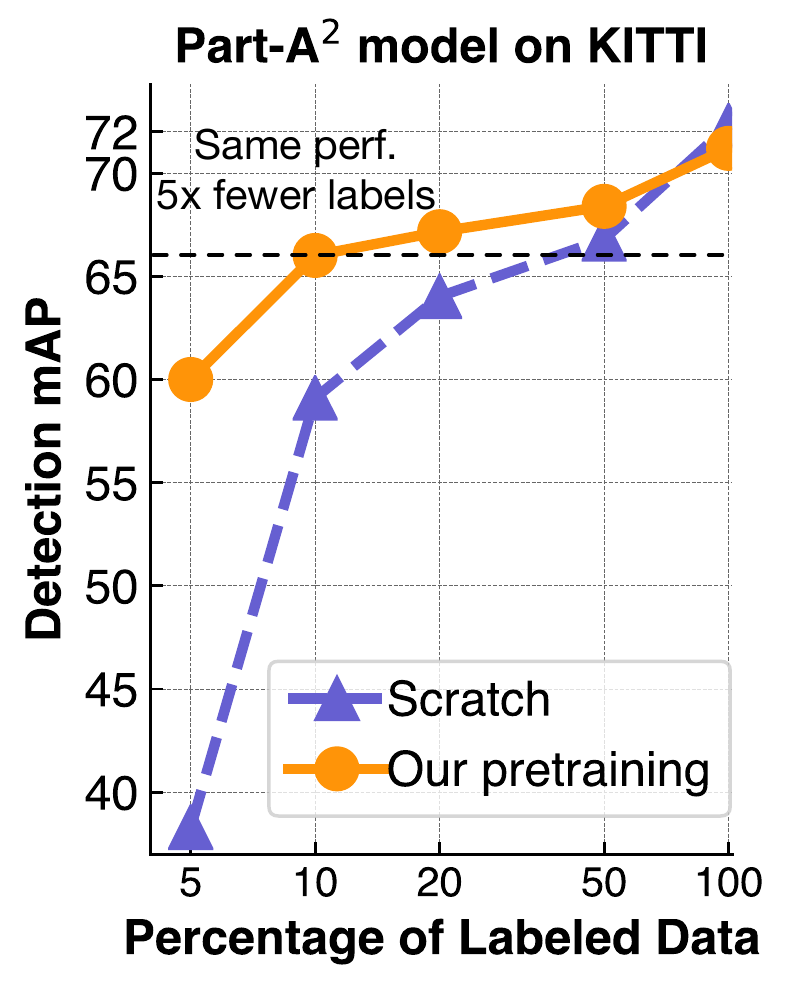}
\vspace{-0.1in}
\caption{\textbf{Using outdoor LiDAR data.}
We finetune detection models from scratch or using our pretraining and report mAP (with 40 recall positions) on the \texttt{cyclist} class at moderate difficulty level of the KITTI val split.
Our models are pretrained using unlabeled outdoor data from the Waymo dataset and outperform scratch training using either point (left) or voxel (right) inputs.
}
\label{fig:lidar}
\end{figure}

\subsection{Generalization to Outdoor LiDAR data}
\label{sec:outdoor}
We test \algorithmName's generalization to outdoor LiDAR data by pretraining on the Waymo Open Dataset~\cite{sun2020scalability} where we extract 79K single-view scans from the videos.
We use the same data augmentation parameters from~\cref{sec:data_aug} and only modify random cuboid to work on the full scale of the $Z$ (depth) dimension of the scene.
We use the standard LiDAR-specific model architectures as our format-specific encoders - PointnetMSG~\cite{shi2019pointrcnn} for point clouds and Spconv-UNet~\cite{shi2019points} for voxels.
Similar to~\cref{sec:implementation_details}, we obtain features from these models after global max pooling and a two layer MLP.
The models are optimized jointly with~\cref{eqn:loss_final} using both within and across-format losses.
For transfer learning, we use the standard KITTI~\cite{Geiger2013IJRR} object detection benchmark, and PointRCNN~\cite{shi2019pointrcnn} and Part-$A^2$~\cite{shi2019points} for down-stream models.
We report results on the \texttt{cyclist} class since it has fewer examples in the training set compared to the other classes.
We provide results for other classes and finetuning details in the supplemental material.
Similar to~\cref{sec:label_efficiency}, while varying the fraction of pretraining data, we report average performance across 3 independent samplings of the data.
~\cref{fig:lidar} shows that our pretrained models outperform training from scratch especially when finetuning on fewer training samples.
For Spconv-Unet, we achieve a \textbf{20\% gain with 5\% of labeled data}.
This suggests that \algorithmName pretraining generalizes across multiple input formats, and our proposed data augmentation generalizes to different depth sensors and scene types.

\vspace{-0.05in}
\section{Conclusion}
\label{sec:discussion}
\vspace{-0.05in}
We propose \algorithmName - an easy to implement self-supervised method that works across model architectures, input data formats, indoor/outdoor 3D, single/multi-view 3D data.
\algorithmName pretrains high capacity models for 3D recognition tasks, and leverages large scale 3D data that may not have multi-view information.
We show state-of-the-art performance on detection and segmentation benchmarks, outperforming all prior work on detection.
We provide crucial insights that make our simple implementation work well - training jointly with multiple input data formats, and designing a generalizable data augmentation scheme.
We hope \algorithmName helps future work in 3D self-supervised learning.

{\small
\bibliographystyle{ieee_fullname}
\bibliography{refs}

\begin{thebibliography}{100}\itemsep=-1pt

\bibitem{achituve2020self}
Idan Achituve, Haggai Maron, and Gal Chechik.
\newblock Self-supervised learning for domain adaptation on point-clouds.
\newblock {\em arXiv preprint arXiv:2003.12641}, 2020.

\bibitem{achlioptas2018learning}
Panos Achlioptas, Olga Diamanti, Ioannis Mitliagkas, and Leonidas Guibas.
\newblock Learning representations and generative models for 3d point clouds.
\newblock In {\em Proceedings of the International Conference on Machine
  Learning (ICML)}, pages 40--49. PMLR, 2018.

\bibitem{adams2010fast}
Andrew Adams, Jongmin Baek, and Myers~Abraham Davis.
\newblock Fast high-dimensional filtering using the permutohedral lattice.
\newblock In {\em Computer Graphics Forum}, volume~29, pages 753--762. Wiley
  Online Library, 2010.

\bibitem{2D-3D-S}
Iro Armeni, Sasha Sax, Amir~Roshan Zamir, and Silvio Savarese.
\newblock Joint 2d-3d-semantic data for indoor scene understanding.
\newblock {\em CoRR}, abs/1702.01105, 2017.

\bibitem{armeni20163d}
Iro Armeni, Ozan Sener, Amir~R Zamir, Helen Jiang, Ioannis Brilakis, Martin
  Fischer, and Silvio Savarese.
\newblock 3d semantic parsing of large-scale indoor spaces.
\newblock In {\em Proceedings of the IEEE Conference on Computer Vision and
  Pattern Recognition}, pages 1534--1543, 2016.

\bibitem{atzmon2018point}
Matan Atzmon, Haggai Maron, and Yaron Lipman.
\newblock Point convolutional neural networks by extension operators.
\newblock {\em arXiv preprint arXiv:1803.10091}, 2018.

\bibitem{bojanowski2017unsupervised}
Piotr Bojanowski and Armand Joulin.
\newblock Unsupervised learning by predicting noise.
\newblock In {\em Proceedings of the International Conference on Machine
  Learning (ICML)}, 2017.

\bibitem{caron2018deep}
Mathilde Caron, Piotr Bojanowski, Armand Joulin, and Matthijs Douze.
\newblock Deep clustering for unsupervised learning of visual features.
\newblock In {\em Proceedings of the European Conference on Computer Vision
  (ECCV)}, 2018.

\bibitem{caron2020swav}
Mathilde Caron, Ishan Misra, Julien Mairal, Priya Goyal, Piotr Bojanowski, and
  Armand Joulin.
\newblock Unsupervised learning of visual features by contrasting cluster
  assignments.
\newblock In {\em NeurIPS}, 2020.

\bibitem{Matterport3D}
Angel Chang, Angela Dai, Thomas Funkhouser, Maciej Halber, Matthias Niessner,
  Manolis Savva, Shuran Song, Andy Zeng, and Yinda Zhang.
\newblock Matterport3d: Learning from rgb-d data in indoor environments.
\newblock {\em International Conference on 3D Vision (3DV)}, 2017.
\newblock Matterport license available at
  http://kaldir.vc.in.tum.de/matterport/MP\_TOS.pdf.

\bibitem{chang2015shapenet}
Angel~X Chang, Thomas Funkhouser, Leonidas Guibas, Pat Hanrahan, Qixing Huang,
  Zimo Li, Silvio Savarese, Manolis Savva, Shuran Song, Hao Su, et~al.
\newblock Shapenet: An information-rich 3d model repository.
\newblock {\em arXiv preprint arXiv:1512.03012}, 2015.

\bibitem{chen2020hierarchical}
Jintai Chen, Biwen Lei, Qingyu Song, Haochao Ying, Danny~Z Chen, and Jian Wu.
\newblock A hierarchical graph network for 3d object detection on point clouds.
\newblock In {\em Proceedings of the IEEE/CVF Conference on Computer Vision and
  Pattern Recognition}, pages 392--401, 2020.

\bibitem{chen2020simple}
Ting Chen, Simon Kornblith, Mohammad Norouzi, and Geoffrey Hinton.
\newblock A simple framework for contrastive learning of visual
  representations.
\newblock {\em arXiv preprint arXiv:2002.05709}, 2020.

\bibitem{chen2020improved}
Xinlei Chen, Haoqi Fan, Ross Girshick, and Kaiming He.
\newblock Improved baselines with momentum contrastive learning.
\newblock {\em arXiv preprint arXiv:2003.04297}, 2020.

\bibitem{chen2019fast}
Yilun Chen, Shu Liu, Xiaoyong Shen, and Jiaya Jia.
\newblock Fast point r-cnn.
\newblock In {\em Proceedings of the IEEE International Conference on Computer
  Vision}, pages 9775--9784, 2019.

\bibitem{Choi2016}
Sungjoon Choi, Qian-Yi Zhou, Stephen Miller, and Vladlen Koltun.
\newblock A large dataset of object scans.
\newblock {\em arXiv:1602.02481}, 2016.

\bibitem{choy20194d}
Christopher Choy, JunYoung Gwak, and Silvio Savarese.
\newblock 4d spatio-temporal convnets: Minkowski convolutional neural networks.
\newblock In {\em Proceedings of the IEEE Conference on Computer Vision and
  Pattern Recognition}, pages 3075--3084, 2019.

\bibitem{Dai_2017_CVPR_scannet}
Angela Dai, Angel~X. Chang, Manolis Savva, Maciej Halber, Thomas Funkhouser,
  and Matthias Niessner.
\newblock Scannet: Richly-annotated 3d reconstructions of indoor scenes.
\newblock In {\em The IEEE Conference on Computer Vision and Pattern
  Recognition (CVPR)}, July 2017.

\bibitem{doersch2015unsupervised}
Carl Doersch, Abhinav Gupta, and Alexei~A Efros.
\newblock Unsupervised visual representation learning by context prediction.
\newblock In {\em Proceedings of the International Conference on Computer
  Vision (ICCV)}, 2015.

\bibitem{donahue2017adversarial}
J. Donahue, P. Krahenb{\"u}hl, and T. Darrell.
\newblock Adversarial feature learning.
\newblock In {\em International Conference on Learning Representations (ICLR)},
  2016.

\bibitem{dosovitskiy2016discriminative}
Alexey Dosovitskiy, Philipp Fischer, Jost~Tobias Springenberg, Martin
  Riedmiller, and Thomas Brox.
\newblock Discriminative unsupervised feature learning with exemplar
  convolutional neural networks.
\newblock {\em IEEE Transactions on Pattern Analysis and Machine Intelligence
  (TPAMI)}, 38(9):1734--1747, 2016.

\bibitem{engelmann20203d}
Francis Engelmann, Martin Bokeloh, Alireza Fathi, Bastian Leibe, and Matthias
  Nie{\ss}ner.
\newblock 3d-mpa: Multi-proposal aggregation for 3d semantic instance
  segmentation.
\newblock In {\em Proceedings of the IEEE/CVF Conference on Computer Vision and
  Pattern Recognition}, pages 9031--9040, 2020.

\bibitem{fan2017point}
Haoqiang Fan, Hao Su, and Leonidas~J Guibas.
\newblock A point set generation network for 3d object reconstruction from a
  single image.
\newblock In {\em Proceedings of the IEEE conference on computer vision and
  pattern recognition}, pages 605--613, 2017.

\bibitem{franklin2020ipad}
Cat Franklin.
\newblock Apple unveils new ipad pro with breakthrough lidar scanner and brings
  trackpad support to ipados.
\newblock \url{https://www.apple.com/}, 2020.

\bibitem{gadelha2018multiresolution}
Matheus Gadelha, Rui Wang, and Subhransu Maji.
\newblock Multiresolution tree networks for 3d point cloud processing.
\newblock In {\em Proceedings of the European Conference on Computer Vision
  (ECCV)}, pages 103--118, 2018.

\bibitem{Geiger2013IJRR}
Andreas Geiger, Philip Lenz, Christoph Stiller, and Raquel Urtasun.
\newblock Vision meets robotics: The kitti dataset.
\newblock {\em International Journal of Robotics Research (IJRR)}, 2013.

\bibitem{ghadiyaram2019large}
Deepti Ghadiyaram, Matt Feiszli, Du Tran, Xueting Yan, Heng Wang, and Dhruv
  Mahajan.
\newblock Large-scale weakly-supervised pre-training for video action
  recognition.
\newblock {\em arXiv preprint arXiv:1905.00561}, 2019.

\bibitem{girdhar16learning}
Rohit Girdhar, David~Ford Fouhey, Mikel Rodriguez, and Abhinav Gupta.
\newblock Learning a predictable and generative vector representation for
  objects.
\newblock In {\em Proceedings of the European Conference on Computer Vision
  (ECCV)}, 2016.

\bibitem{gkioxari2019mesh}
Georgia Gkioxari, Jitendra Malik, and Justin Johnson.
\newblock Mesh r-cnn.
\newblock In {\em Proceedings of the IEEE International Conference on Computer
  Vision}, pages 9785--9795, 2019.

\bibitem{goyal2019scaling}
Priya Goyal, Dhruv Mahajan, Abhinav Gupta, and Ishan Misra.
\newblock Scaling and benchmarking self-supervised visual representation
  learning.
\newblock {\em Proceedings of the International Conference on Computer Vision
  (ICCV)}, 2019.

\bibitem{graham2015sparse}
Ben Graham.
\newblock Sparse 3d convolutional neural networks.
\newblock {\em arXiv preprint arXiv:1505.02890}, 2015.

\bibitem{graham20183d}
Benjamin Graham, Martin Engelcke, and Laurens van~der Maaten.
\newblock 3d semantic segmentation with submanifold sparse convolutional
  networks.
\newblock In {\em Proceedings of the IEEE Conference on Computer Vision and
  Pattern Recognition}, pages 9224--9232, 2018.

\bibitem{groueix2018papier}
Thibault Groueix, Matthew Fisher, Vladimir~G Kim, Bryan~C Russell, and Mathieu
  Aubry.
\newblock A papier-m{\^a}ch{\'e} approach to learning 3d surface generation.
\newblock In {\em Proceedings of the Conference on Computer Vision and Pattern
  Recognition (CVPR)}, pages 216--224, 2018.

\bibitem{hadsell2006dimensionality}
Raia Hadsell, Sumit Chopra, and Yann LeCun.
\newblock Dimensionality reduction by learning an invariant mapping.
\newblock In {\em Proceedings of the Conference on Computer Vision and Pattern
  Recognition (CVPR)}, 2006.

\bibitem{hassani2019unsupervised}
Kaveh Hassani and Mike Haley.
\newblock Unsupervised multi-task feature learning on point clouds.
\newblock In {\em Proceedings of the International Conference on Computer
  Vision (ICCV)}, pages 8160--8171, 2019.

\bibitem{he2019momentum}
Kaiming He, Haoqi Fan, Yuxin Wu, Saining Xie, and Ross Girshick.
\newblock Momentum contrast for unsupervised visual representation learning.
\newblock {\em arXiv preprint arXiv:1911.05722}, 2019.

\bibitem{henaff2019data}
Olivier~J H{\'e}naff, Ali Razavi, Carl Doersch, SM Eslami, and Aaron van~den
  Oord.
\newblock Data-efficient image recognition with contrastive predictive coding.
\newblock {\em arXiv preprint arXiv:1905.09272}, 2019.

\bibitem{hermosilla2018monte}
Pedro Hermosilla, Tobias Ritschel, Pere-Pau V{\'a}zquez, {\`A}lvar Vinacua, and
  Timo Ropinski.
\newblock Monte carlo convolution for learning on non-uniformly sampled point
  clouds.
\newblock {\em ACM Transactions on Graphics (TOG)}, 37(6):1--12, 2018.

\bibitem{scenenn-3dv16}
Binh-Son Hua, Quang-Hieu Pham, Duc~Thanh Nguyen, Minh-Khoi Tran, Lap-Fai Yu,
  and Sai-Kit Yeung.
\newblock Scenenn: A scene meshes dataset with annotations.
\newblock In {\em International Conference on 3D Vision (3DV)}, 2016.

\bibitem{hua2018pointwise}
Binh-Son Hua, Minh-Khoi Tran, and Sai-Kit Yeung.
\newblock Pointwise convolutional neural networks.
\newblock In {\em Proceedings of the IEEE Conference on Computer Vision and
  Pattern Recognition}, pages 984--993, 2018.

\bibitem{jabri2020space}
Allan Jabri, Andrew Owens, and Alexei~A Efros.
\newblock Space-time correspondence as a contrastive random walk.
\newblock {\em arXiv preprint arXiv:2006.14613}, 2020.

\bibitem{janoch2013category}
Allison Janoch, Sergey Karayev, Yangqing Jia, Jonathan~T Barron, Mario Fritz,
  Kate Saenko, and Trevor Darrell.
\newblock A category-level 3d object dataset: Putting the kinect to work.
\newblock In {\em Consumer depth cameras for computer vision}, pages 141--165.
  Springer, 2013.

\bibitem{ji2019invariant}
Xu Ji, Jo{\~a}o~F Henriques, and Andrea Vedaldi.
\newblock Invariant information clustering for unsupervised image
  classification and segmentation.
\newblock In {\em Proceedings of the International Conference on Computer
  Vision (ICCV)}, 2019.

\bibitem{jing2020self}
Longlong Jing, Yucheng Chen, Ling Zhang, Mingyi He, and Yingli Tian.
\newblock Self-supervised modal and view invariant feature learning.
\newblock {\em arXiv preprint arXiv:2005.14169}, 2020.

\bibitem{joulin2016learning}
Armand Joulin, Laurens Van Der~Maaten, Allan Jabri, and Nicolas Vasilache.
\newblock Learning visual features from large weakly supervised data.
\newblock In {\em European Conference on Computer Vision}, pages 67--84.
  Springer, 2016.

\bibitem{KingmaB14}
Diederik~P. Kingma and Jimmy Ba.
\newblock Adam: A method for stochastic optimization.
\newblock In {\em 3rd International Conference on Learning Representations,
  ICLR 2015, San Diego, CA, USA, May 7-9, 2015, Conference Track Proceedings},
  2015.

\bibitem{klokov2017escape}
Roman Klokov and Victor Lempitsky.
\newblock Escape from cells: Deep kd-networks for the recognition of 3d point
  cloud models.
\newblock In {\em Proceedings of the IEEE International Conference on Computer
  Vision}, pages 863--872, 2017.

\bibitem{kolesnikov2019big}
Alexander Kolesnikov, Lucas Beyer, Xiaohua Zhai, Joan Puigcerver, Jessica Yung,
  Sylvain Gelly, and Neil Houlsby.
\newblock Big transfer (bit): General visual representation learning.
\newblock {\em arXiv preprint arXiv:1912.11370}, 6, 2019.

\bibitem{kolesnikov2019revisiting}
Alexander Kolesnikov, Xiaohua Zhai, and Lucas Beyer.
\newblock Revisiting self-supervised visual representation learning.
\newblock {\em arXiv preprint arXiv:1901.09005}, 2019.

\bibitem{li2019gs3d}
Buyu Li, Wanli Ouyang, Lu Sheng, Xingyu Zeng, and Xiaogang Wang.
\newblock Gs3d: An efficient 3d object detection framework for autonomous
  driving.
\newblock In {\em Proceedings of the IEEE Conference on Computer Vision and
  Pattern Recognition}, pages 1019--1028, 2019.

\bibitem{li2018so}
Jiaxin Li, Ben~M Chen, and Gim Hee~Lee.
\newblock So-net: Self-organizing network for point cloud analysis.
\newblock In {\em Proceedings of the Conference on Computer Vision and Pattern
  Recognition (CVPR)}, pages 9397--9406, 2018.

\bibitem{li2018pointcnn}
Yangyan Li, Rui Bu, Mingchao Sun, Wei Wu, Xinhan Di, and Baoquan Chen.
\newblock Pointcnn: Convolution on x-transformed points.
\newblock In {\em Advances in Neural Information Processing Systems}, pages
  820--830, 2018.

\bibitem{loshchilov2016sgdr}
Ilya Loshchilov and Frank Hutter.
\newblock Sgdr: Stochastic gradient descent with warm restarts.
\newblock {\em arXiv preprint arXiv:1608.03983}, 2016.

\bibitem{loshchilov2017decoupled}
Ilya Loshchilov and Frank Hutter.
\newblock Decoupled weight decay regularization.
\newblock {\em arXiv preprint arXiv:1711.05101}, 2017.

\bibitem{mahajan2018exploring}
Dhruv Mahajan, Ross Girshick, Vignesh Ramanathan, Kaiming He, Manohar Paluri,
  Yixuan Li, Ashwin Bharambe, and Laurens van~der Maaten.
\newblock Exploring the limits of weakly supervised pretraining.
\newblock In {\em ECCV}, 2018.

\bibitem{masci2011stacked}
J. Masci, U. Meier, D. Cires, and J. Schmidhuber.
\newblock Stacked convolutional auto-encoders for hierarchical feature
  extraction.
\newblock In {\em ICANN}, pages 52--59, 2011.

\bibitem{maturana2015voxnet}
Daniel Maturana and Sebastian Scherer.
\newblock Voxnet: A 3d convolutional neural network for real-time object
  recognition.
\newblock In {\em 2015 IEEE/RSJ International Conference on Intelligent Robots
  and Systems (IROS)}, pages 922--928. IEEE, 2015.

\bibitem{mescheder2017unifying}
L. Mescheder, S. Nowozin, and A. Geiger.
\newblock Adversarial variational {B}ayes: Unifying variational autoencoders
  and generative adversarial networks.
\newblock In {\em Proceedings of the International Conference on Machine
  Learning (ICML)}, 2017.

\bibitem{misra2019self}
Ishan Misra and Laurens van~der Maaten.
\newblock Self-supervised learning of pretext-invariant representations.
\newblock {\em arXiv preprint arXiv:1912.01991}, 2019.

\bibitem{morgado2020avid}
Pedro Morgado, Nuno Vasconcelos, and Ishan Misra.
\newblock Audio-visual instance discrimination with cross-modal agreement.
\newblock {\em https://arxiv.org/abs/2004.12943}, 2020.

\bibitem{Silberman_ECCV12_NYUv2}
Pushmeet~Kohli Nathan~Silberman, Derek~Hoiem and Rob Fergus.
\newblock Indoor segmentation and support inference from rgbd images.
\newblock In {\em Proceedings of the European Conference on Computer Vision
  (ECCV)}, 2012.

\bibitem{noroozi2016unsupervised}
Mehdi Noroozi and Paolo Favaro.
\newblock Unsupervised learning of visual representations by solving jigsaw
  puzzles.
\newblock In {\em Proceedings of the European Conference on Computer Vision
  (ECCV)}, 2016.

\bibitem{olshausen1996}
B.~A. Olshausen and D.~J. Field.
\newblock Emergence of simple-cell receptive field properties by learning a
  sparse code for natural images.
\newblock {\em Nature}, 381(6583):607, 1996.

\bibitem{oord2018representation}
Aaron van~den Oord, Yazhe Li, and Oriol Vinyals.
\newblock Representation learning with contrastive predictive coding.
\newblock {\em arXiv preprint arXiv:1807.03748}, 2018.

\bibitem{patrick2020multi}
Mandela Patrick, Yuki~M Asano, Ruth Fong, Jo{\~a}o~F Henriques, Geoffrey Zweig,
  and Andrea Vedaldi.
\newblock Multi-modal self-supervision from generalized data transformations.
\newblock {\em arXiv preprint arXiv:2003.04298}, 2020.

\bibitem{pham2019jsis3d}
Quang-Hieu Pham, Thanh Nguyen, Binh-Son Hua, Gemma Roig, and Sai-Kit Yeung.
\newblock Jsis3d: Joint semantic-instance segmentation of 3d point clouds with
  multi-task pointwise networks and multi-value conditional random fields.
\newblock In {\em Proceedings of the IEEE Conference on Computer Vision and
  Pattern Recognition}, pages 8827--8836, 2019.

\bibitem{qi2019votenet}
Charles~R Qi, Or Litany, Kaiming He, and Leonidas~J Guibas.
\newblock Deep hough voting for 3d object detection in point clouds.
\newblock {\em arXiv preprint arXiv:1904.09664}, 2019.

\bibitem{qi2018frustum}
Charles~R Qi, Wei Liu, Chenxia Wu, Hao Su, and Leonidas~J Guibas.
\newblock Frustum pointnets for 3d object detection from rgb-d data.
\newblock In {\em Proceedings of the IEEE Conference on Computer Vision and
  Pattern Recognition}, pages 918--927, 2018.

\bibitem{qi2017pointnet}
Charles~R Qi, Hao Su, Kaichun Mo, and Leonidas~J Guibas.
\newblock Pointnet: Deep learning on point sets for 3d classification and
  segmentation.
\newblock In {\em Proceedings of the IEEE Conference on Computer Vision and
  Pattern Recognition}, pages 652--660, 2017.

\bibitem{qi2016volumetric}
Charles~R Qi, Hao Su, Matthias Nie{\ss}ner, Angela Dai, Mengyuan Yan, and
  Leonidas~J Guibas.
\newblock Volumetric and multi-view cnns for object classification on 3d data.
\newblock In {\em Proceedings of the IEEE conference on computer vision and
  pattern recognition}, pages 5648--5656, 2016.

\bibitem{qi2017pointnet++}
Charles~Ruizhongtai Qi, Li Yi, Hao Su, and Leonidas~J Guibas.
\newblock Pointnet++: Deep hierarchical feature learning on point sets in a
  metric space.
\newblock In {\em Advances in neural information processing systems}, pages
  5099--5108, 2017.

\bibitem{ranzato2007unsupervised}
Marc’Aurelio Ranzato, Fu-Jie Huang, Y-Lan Boureau, and Yann LeCun.
\newblock Unsupervised learning of invariant feature hierarchies with
  applications to object recognition.
\newblock In {\em CVPR}, 2007.

\bibitem{riegler2017octnet}
Gernot Riegler, Ali Osman~Ulusoy, and Andreas Geiger.
\newblock Octnet: Learning deep 3d representations at high resolutions.
\newblock In {\em Proceedings of the IEEE Conference on Computer Vision and
  Pattern Recognition}, pages 3577--3586, 2017.

\bibitem{RosCVPR16}
German Ros, Laura Sellart, Joanna Materzynska, David Vazquez, and Antonio
  Lopez.
\newblock {The SYNTHIA Dataset}: A large collection of synthetic images for
  semantic segmentation of urban scenes.
\newblock In {\em Proceedings of the Conference on Computer Vision and Pattern
  Recognition (CVPR)}, 2016.

\bibitem{salakhutdinov2009deep}
R. Salakhutdinov and G. Hinton.
\newblock Deep {B}oltzmann machines.
\newblock In {\em AI-STATS}, pages 448--455, 2009.

\bibitem{samsung-depth-sensor}
Samsung.
\newblock What is depthvision camera on galaxy s20+ and s20 ultra?
\newblock \url{https://www.samsung.com/}, 2020.

\bibitem{sauder2019self}
Jonathan Sauder and Bjarne Sievers.
\newblock Self-supervised deep learning on point clouds by reconstructing
  space.
\newblock In {\em Advances in Neural Information Processing Systems (NeurIPS)},
  pages 12962--12972, 2019.

\bibitem{savva2016pigraphs}
Manolis Savva, Angel~X Chang, Pat Hanrahan, Matthew Fisher, and Matthias
  Nie{\ss}ner.
\newblock Pigraphs: learning interaction snapshots from observations.
\newblock {\em ACM Transactions on Graphics (TOG)}, 35(4):1--12, 2016.

\bibitem{shi2020pv}
Shaoshuai Shi, Chaoxu Guo, Li Jiang, Zhe Wang, Jianping Shi, Xiaogang Wang, and
  Hongsheng Li.
\newblock Pv-rcnn: Point-voxel feature set abstraction for 3d object detection.
\newblock In {\em Proceedings of the IEEE/CVF Conference on Computer Vision and
  Pattern Recognition}, pages 10529--10538, 2020.

\bibitem{shi2019pointrcnn}
Shaoshuai Shi, Xiaogang Wang, and Hongsheng Li.
\newblock Pointrcnn: 3d object proposal generation and detection from point
  cloud.
\newblock In {\em Proceedings of the IEEE Conference on Computer Vision and
  Pattern Recognition}, pages 770--779, 2019.

\bibitem{shi2019points}
Shaoshuai Shi, Zhe Wang, Jianping Shi, Xiaogang Wang, and Hongsheng Li.
\newblock From points to parts: 3d object detection from point cloud with
  part-aware and part-aggregation network.
\newblock {\em arXiv preprint arXiv:1907.03670}, 2019.

\bibitem{silberman2012indoor}
Nathan Silberman, Derek Hoiem, Pushmeet Kohli, and Rob Fergus.
\newblock Indoor segmentation and support inference from rgbd images.
\newblock In {\em European conference on computer vision}, pages 746--760.
  Springer, 2012.

\bibitem{sohn2016improved}
Kihyuk Sohn.
\newblock Improved deep metric learning with multi-class n-pair loss objective.
\newblock In {\em Advances in Neural Information Processing Systems (NeurIPS)},
  2016.

\bibitem{song2015sun}
Shuran Song, Samuel~P Lichtenberg, and Jianxiong Xiao.
\newblock Sun rgb-d: A rgb-d scene understanding benchmark suite.
\newblock In {\em Proceedings of the IEEE conference on computer vision and
  pattern recognition}, pages 567--576, 2015.

\bibitem{song2016deep}
Shuran Song and Jianxiong Xiao.
\newblock Deep sliding shapes for amodal 3d object detection in rgb-d images.
\newblock In {\em Proceedings of the IEEE Conference on Computer Vision and
  Pattern Recognition}, pages 808--816, 2016.

\bibitem{stein2020iphone}
Scott Stein.
\newblock Lidar on the iphone 12 pro.
\newblock \url{https://www.cnet.com/}, 2020.

\bibitem{sturm2012benchmark}
J{\"u}rgen Sturm, Nikolas Engelhard, Felix Endres, Wolfram Burgard, and Daniel
  Cremers.
\newblock A benchmark for the evaluation of rgb-d slam systems.
\newblock In {\em 2012 IEEE/RSJ International Conference on Intelligent Robots
  and Systems}, pages 573--580. IEEE, 2012.

\bibitem{su2018splatnet}
Hang Su, Varun Jampani, Deqing Sun, Subhransu Maji, Evangelos Kalogerakis,
  Ming-Hsuan Yang, and Jan Kautz.
\newblock Splatnet: Sparse lattice networks for point cloud processing.
\newblock In {\em Proceedings of the IEEE Conference on Computer Vision and
  Pattern Recognition}, pages 2530--2539, 2018.

\bibitem{su2015multi}
Hang Su, Subhransu Maji, Evangelos Kalogerakis, and Erik Learned-Miller.
\newblock Multi-view convolutional neural networks for 3d shape recognition.
\newblock In {\em Proceedings of the IEEE international conference on computer
  vision}, pages 945--953, 2015.

\bibitem{sun2017revisiting}
Chen Sun, Abhinav Shrivastava, Saurabh Singh, and Abhinav Gupta.
\newblock Revisiting unreasonable effectiveness of data in deep learning era.
\newblock In {\em ICCV}, 2017.

\bibitem{sun2020scalability}
Pei Sun, Henrik Kretzschmar, Xerxes Dotiwalla, Aurelien Chouard, Vijaysai
  Patnaik, Paul Tsui, James Guo, Yin Zhou, Yuning Chai, Benjamin Caine, et~al.
\newblock Scalability in perception for autonomous driving: Waymo open dataset.
\newblock In {\em Proceedings of the IEEE/CVF Conference on Computer Vision and
  Pattern Recognition}, pages 2446--2454, 2020.
\newblock This publication was made using the Waymo Open Dataset, provided by
  Waymo LLC under license terms available at waymo.com/open.

\bibitem{szegedy2015going}
Christian Szegedy, Wei Liu, Yangqing Jia, Pierre Sermanet, Scott Reed, Dragomir
  Anguelov, Dumitru Erhan, Vincent Vanhoucke, and Andrew Rabinovich.
\newblock Going deeper with convolutions.
\newblock In {\em Proceedings of the Conference on Computer Vision and Pattern
  Recognition (CVPR)}, 2015.

\bibitem{tatarchenko2017octree}
Maxim Tatarchenko, Alexey Dosovitskiy, and Thomas Brox.
\newblock Octree generating networks: Efficient convolutional architectures for
  high-resolution 3d outputs.
\newblock In {\em Proceedings of the IEEE International Conference on Computer
  Vision}, pages 2088--2096, 2017.

\bibitem{tatarchenko2018tangent}
Maxim Tatarchenko, Jaesik Park, Vladlen Koltun, and Qian-Yi Zhou.
\newblock Tangent convolutions for dense prediction in 3d.
\newblock In {\em Proceedings of the IEEE Conference on Computer Vision and
  Pattern Recognition}, pages 3887--3896, 2018.

\bibitem{tchapmi2017segcloud}
Lyne Tchapmi, Christopher Choy, Iro Armeni, JunYoung Gwak, and Silvio Savarese.
\newblock Segcloud: Semantic segmentation of 3d point clouds.
\newblock In {\em 2017 international conference on 3D vision (3DV)}, pages
  537--547. IEEE, 2017.

\bibitem{openpcdet2020}
OpenPCDet~Development Team.
\newblock Openpcdet: An open-source toolbox for 3d object detection from point
  clouds.
\newblock \url{https://github.com/open-mmlab/OpenPCDet}, 2020.

\bibitem{tian2019contrastive}
Yonglong Tian, Dilip Krishnan, and Phillip Isola.
\newblock Contrastive multiview coding.
\newblock {\em arXiv preprint arXiv:1906.05849}, 2019.

\bibitem{tian2020makes}
Yonglong Tian, Chen Sun, Ben Poole, Dilip Krishnan, Cordelia Schmid, and
  Phillip Isola.
\newblock What makes for good views for contrastive learning?
\newblock {\em arXiv preprint arXiv:2005.10243}, 2020.

\bibitem{vincent2008extracting}
P. Vincent, H. Larochelle, Y. Bengio, and P.-A. Manzagol.
\newblock Extracting and composing robust features with denoising autoencoders.
\newblock In {\em ICML}, 2008.

\bibitem{wang2018pixel2mesh}
Nanyang Wang, Yinda Zhang, Zhuwen Li, Yanwei Fu, Wei Liu, and Yu-Gang Jiang.
\newblock Pixel2mesh: Generating 3d mesh models from single rgb images.
\newblock In {\em Proceedings of the European Conference on Computer Vision
  (ECCV)}, pages 52--67, 2018.

\bibitem{wang2015unsupervised}
Xiaolong Wang and Abhinav Gupta.
\newblock Unsupervised learning of visual representations using videos.
\newblock In {\em Proceedings of the International Conference on Computer
  Vision (ICCV)}, 2015.

\bibitem{wang2019associatively}
Xinlong Wang, Shu Liu, Xiaoyong Shen, Chunhua Shen, and Jiaya Jia.
\newblock Associatively segmenting instances and semantics in point clouds.
\newblock In {\em Proceedings of the IEEE Conference on Computer Vision and
  Pattern Recognition}, pages 4096--4105, 2019.

\bibitem{wang2019deep}
Yue Wang and Justin~M Solomon.
\newblock Deep closest point: Learning representations for point cloud
  registration.
\newblock In {\em Proceedings of the International Conference on Computer
  Vision (ICCV)}, pages 3523--3532, 2019.

\bibitem{wang2018dynamic}
Yue Wang, Yongbin Sun, Ziwei Liu, Sanjay~E Sarma, Michael~M Bronstein, and
  Justin~M Solomon.
\newblock Dynamic graph cnn for learning on point clouds.
\newblock {\em arXiv preprint arXiv:1801.07829}, 2018.

\bibitem{wu2019pointconv}
Wenxuan Wu, Zhongang Qi, and Li Fuxin.
\newblock Pointconv: Deep convolutional networks on 3d point clouds.
\newblock In {\em Proceedings of the IEEE Conference on Computer Vision and
  Pattern Recognition}, pages 9621--9630, 2019.

\bibitem{WuSKYZTX15}
Zhirong Wu, Shuran Song, Aditya Khosla, Fisher Yu, Linguang Zhang, Xiaoou Tang,
  and Jianxiong Xiao.
\newblock 3d shapenets: {A} deep representation for volumetric shapes.
\newblock In {\em {IEEE} Conference on Computer Vision and Pattern Recognition,
  {CVPR} 2015, Boston, MA, USA, June 7-12, 2015}, pages 1912--1920, 2015.

\bibitem{wu2018unsupervised}
Zhirong Wu, Yuanjun Xiong, Stella~X Yu, and Dahua Lin.
\newblock Unsupervised feature learning via non-parametric instance
  discrimination.
\newblock In {\em Proceedings of the Conference on Computer Vision and Pattern
  Recognition (CVPR)}, 2018.

\bibitem{xiao2013sun3d}
Jianxiong Xiao, Andrew Owens, and Antonio Torralba.
\newblock Sun3d: A database of big spaces reconstructed using sfm and object
  labels.
\newblock In {\em Proceedings of the IEEE international conference on computer
  vision}, pages 1625--1632, 2013.

\bibitem{xie2020pointcontrast}
Saining Xie, Jiatao Gu, Demi Guo, Charles~R Qi, Leonidas~J Guibas, and Or
  Litany.
\newblock Pointcontrast: Unsupervised pre-training for 3d point cloud
  understanding.
\newblock In {\em Proceedings of the European Conference on Computer Vision
  (ECCV)}, 2020.

\bibitem{xie2018attentional}
Saining Xie, Sainan Liu, Zeyu Chen, and Zhuowen Tu.
\newblock Attentional shapecontextnet for point cloud recognition.
\newblock In {\em Proceedings of the IEEE Conference on Computer Vision and
  Pattern Recognition}, pages 4606--4615, 2018.

\bibitem{xu2018spidercnn}
Yifan Xu, Tianqi Fan, Mingye Xu, Long Zeng, and Yu Qiao.
\newblock Spidercnn: Deep learning on point sets with parameterized
  convolutional filters.
\newblock In {\em Proceedings of the European Conference on Computer Vision
  (ECCV)}, pages 87--102, 2018.

\bibitem{yan2020pointasnl}
Xu Yan, Chaoda Zheng, Zhen Li, Sheng Wang, and Shuguang Cui.
\newblock Pointasnl: Robust point clouds processing using nonlocal neural
  networks with adaptive sampling.
\newblock In {\em Proceedings of the IEEE/CVF Conference on Computer Vision and
  Pattern Recognition}, pages 5589--5598, 2020.

\bibitem{yan2018second}
Yan Yan, Yuxing Mao, and Bo Li.
\newblock Second: Sparsely embedded convolutional detection.
\newblock {\em Sensors}, 18(10):3337, 2018.

\bibitem{yang2018foldingnet}
Yaoqing Yang, Chen Feng, Yiru Shen, and Dong Tian.
\newblock Foldingnet: Point cloud auto-encoder via deep grid deformation.
\newblock In {\em Proceedings of the Conference on Computer Vision and Pattern
  Recognition (CVPR)}, pages 206--215, 2018.

\bibitem{yang2019std}
Zetong Yang, Yanan Sun, Shu Liu, Xiaoyong Shen, and Jiaya Jia.
\newblock Std: Sparse-to-dense 3d object detector for point cloud.
\newblock In {\em Proceedings of the IEEE International Conference on Computer
  Vision}, pages 1951--1960, 2019.

\bibitem{yi2019gspn}
Li Yi, Wang Zhao, He Wang, Minhyuk Sung, and Leonidas~J Guibas.
\newblock Gspn: Generative shape proposal network for 3d instance segmentation
  in point cloud.
\newblock In {\em Proceedings of the IEEE Conference on Computer Vision and
  Pattern Recognition}, pages 3947--3956, 2019.

\bibitem{zhang2019path}
Zaiwei Zhang, Zhenxiao Liang, Lemeng Wu, Xiaowei Zhou, and Qixing Huang.
\newblock Path-invariant map networks.
\newblock In {\em Proceedings of the IEEE Conference on Computer Vision and
  Pattern Recognition}, pages 11084--11094, 2019.

\bibitem{zhang2020h3dnet}
Zaiwei Zhang, Bo Sun, Haitao Yang, and Qixing Huang.
\newblock H3dnet: 3d object detection using hybrid geometric primitives.
\newblock {\em arXiv preprint arXiv:2006.05682}, 2020.

\bibitem{zhang2020deep}
Zaiwei Zhang, Zhenpei Yang, Chongyang Ma, Linjie Luo, Alexander Huth, Etienne
  Vouga, and Qixing Huang.
\newblock Deep generative modeling for scene synthesis via hybrid
  representations.
\newblock {\em ACM Transactions on Graphics (TOG)}, 39(2):1--21, 2020.

\bibitem{zhou2018voxelnet}
Yin Zhou and Oncel Tuzel.
\newblock Voxelnet: End-to-end learning for point cloud based 3d object
  detection.
\newblock In {\em Proceedings of the IEEE Conference on Computer Vision and
  Pattern Recognition}, pages 4490--4499, 2018.

\end{thebibliography}
}

\clearpage
\appendix
\section*{Supplemental Material}

We provide the model architecture details in~\cref{supp_sec:arch_detail}.
Hyper-parameters used in training and fine-tuning for \pointnetplus and \mink and additional results are shown in~\cref{supp_sec:train_detail}. We also show hyper-parameters used in training and fine-tuning for PointnetMSG and Spconv-UNet and results for other categories of detection in KITTI in~\cref{supp_sec:lidar_detail}.

\section{Architecture Details}
\label{supp_sec:arch_detail}

\begin{table}[b!]
 	\centering
 	\setlength{\tabcolsep}{0.2em}\resizebox{\linewidth}{!}{
 	\begin{tabular}{| c | c | c | c | c |}
 		\hline
 		layer name & input layer & type & output size & layer params \\
 		\hline
 		sa1 & point cloud (xyz) & SA & (2048,3+128) & (2048,0.2,[64, 64, 128]) \\
 		sa2 & sa1 & SA & (1024,3+256) & (1024,0.4,[128, 128, 256]) \\
 		sa3 & sa2 & SA & (512,3+256) & (512,0.8,[128, 128, 256]) \\
 		sa4 & sa3 & SA & (256,3+256) & (256,1.2,[128, 128, 256]) \\
 		fp1 & sa3,sa4 & FP & (512,3+512) & [512, 512] \\
 		fp2 & sa2,sa3 & FP & (1024,3+512) & [512, 512] \\
 		\hline
 	\end{tabular}
	}
 	\caption{\textbf{\pointnetplus Network Architecture} used in ~\cref{sec:experiments,sec:analyze_data_aug,sec:analyze_single_view}}
 	\label{Table:Arch:Pointnet}
\end{table}

\begin{table}[b!]
	\centering
	\setlength{\tabcolsep}{0.2em}\resizebox{\linewidth}{!}{
		\begin{tabular}{| c | c | c | c | c |}
			\hline
			layer name & input layer & type & output size & layer params \\
			\hline
			sa1 & point cloud (xyz) & SA & (4096,3+96) & (0.1, [16, 16, 32], 0.5, [32, 32, 64]) \\
			sa2 & sa1 & SA & (1024,3+256) & (0.5, [64, 64, 128], 1.0, [64, 96, 128]) \\
			sa3 & sa2 & SA & (256,3+512) & (0.1, [128, 196, 256], 0.5, [128, 196, 256]) \\
			sa4 & sa3 & SA & (64,3+1024) & (0.1, [256, 256, 512], 0.5, [256, 384, 512]) \\
			fp1 & sa3,sa4 & FP & (256,3+1024) & [512, 512] \\
			fp2 & sa2,sa3 & FP & (1024,3+1024) & [512, 512] \\
			fp3 & sa1,sa2 & FP & (4096,3+512) & [256, 256] \\
			fp4 & point cloud,sa1 & FP & (16384,3+256) & [128, 128] \\
			\hline
		\end{tabular}
	}
	\caption{\textbf{PointnetMSG Network Architecture} used in~\cref{sec:outdoor}.}
	\label{Table:Arch:PointnetMSG}
\end{table}

\par \noindent \textbf{\pointnetplus model } used in ~\cref{sec:experiments,sec:analyze_data_aug,sec:analyze_single_view}. As shown in Table~\ref{Table:Arch:Pointnet}, \pointnetplus contains four set abstraction layers and two feature up-sampling layers, designed in~\cite{qi2019votenet}. Each SA layer is specified by $(n, r, [c_1, ..., c_k])$, where $n$ represents number of output points, $r$ represents the ball-region radius of the reception field, $c_i$ represents the feature channel size of the i-th layer in the MLP. Each feature up-sampling (FP) layer upsamples the
point features by interpolating the features on input points
to output points, as designed in~\cite{qi2017pointnet++}. Each FP layer is specified by $[c_1, ..., c_k]$ where $c_i$
is the output of the i-th layer in the MLP. 
In~\cref{sec:larger_models} of the main paper, we create higher capacity versions of the \pointnetplus model by increasing the channel width. For \pointnetplus $2\times$, $3\times$ and $4\times$, we multiply the feature size  by $\{2, 3, 4\}$ of each layer in the MLP of the four set abstraction layers. When applying \pointnetplus in VoteNet and H3DNet, we adjust the rest of the model accordingly based on different point feature size.

\begin{figure}[b!]
	\centering
	\includegraphics[width=0.5\textwidth]{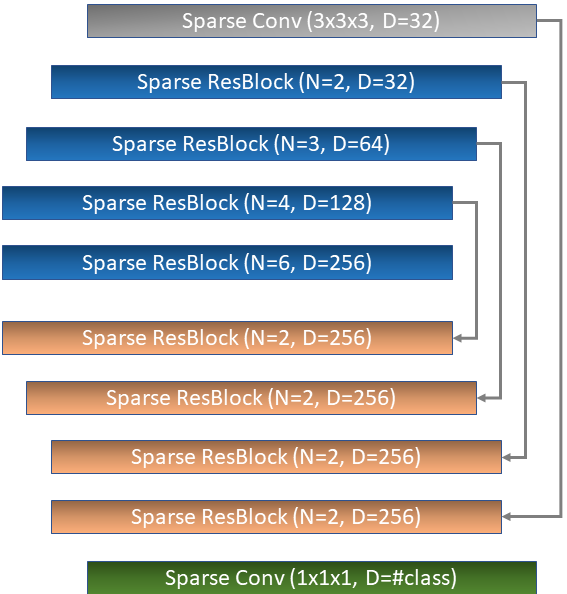}
	\caption{\textbf{\mink Network Architecture} used in~\cref{sec:multiple_input_formats}.}
	\label{fig:unet_arch}
\end{figure}

\vspace{0.05in}
\par \noindent \textbf{PointnetMSG model on LiDAR data} used in~\cref{sec:outdoor}. Table~\ref{Table:Arch:PointnetMSG} shows the architecture details for PointnetMSG, which processes lidar point cloud for PointRCNN detection model. PointnetMSG contains four multi-scale set abstraction layers and four feature up-sampling layers. Multi-scale set abstraction samples points in different scales and process them with different MLPs. In here, we adopt the architecture design in~\cite{shi2019pointrcnn}, in which each set abstraction layer contains two point features produced with two ball-region radius and two MLPs. As shown in~\cref{Table:Arch:PointnetMSG}, each SA layer is specified by $(r^1, [c^1_1, ..., c^1_k], r^2, [c^2_1, ..., c^2_k])$, where $r^1, r^2$ indicates the ball-region radius for each scale and $c^1_i, c^2_i$ indicates the feature channel size of the i-th layer in the MLP of each scale. We directly apply the learnt PointnetMSG in PointRCNN for detection evaluation in KITTI.
\vspace{0.05in}
\par \noindent \textbf{\mink model} used in~\cref{sec:multiple_input_formats}.~\cref{fig:unet_arch} shows the network architecture for \mink. It mainly contains four encoding resblock and four decoding resblock. We use sparse convolution and sparse resblock designed in~\cite{choy20194d}. For each sparse resblock, we first apply sparse convolution or deconvolution, depending on encoding or decoding, with kernel size 2 and stride 2. Then, we apply N number of sparse convolution layers with kernel size 3 and stride 1. D represents the output feature dimension. Since sparse convolution takes variable sized input and output, we do not specify the number of voxels in each layer here. We directly apply the learnt \mink backbone for different scene segmentation tasks.

\begin{figure}[t!]
	\centering
	\includegraphics[width=0.5\textwidth]{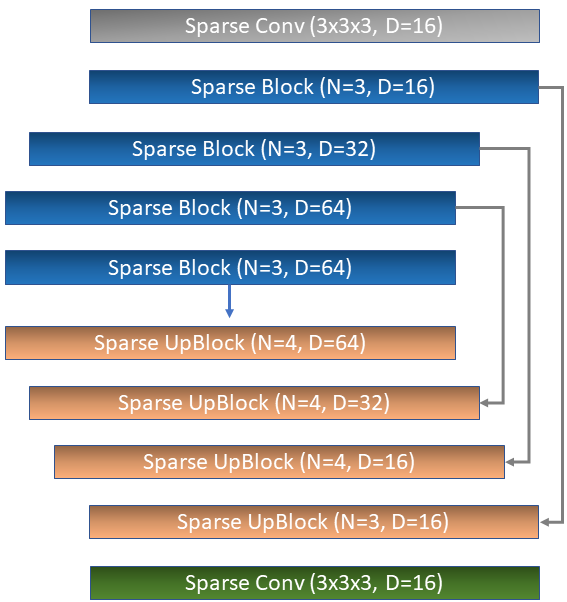}
	\caption{\textbf{Spconv-UNet Network Architecture} used in Section 5.3 of the main paper.}
	\label{fig:unet_arch_lidar}
\end{figure}

\vspace{0.05in}
\par \noindent \textbf{Spconv-UNet model on LiDAR data} used in~\cref{sec:outdoor}.~\cref{fig:unet_arch_lidar} shows the network architecture for Spconv-UNet used for Part $A^2$ detection model. It mainly contains four sparse blocks for encoding and four sparse upblocks for decoding. We use sparse convolution and sparse resblock designed in~\cite{shi2019points}. For each sparse block/upblock, we show the number of convolution layers N and output feature dimension D. We directly adopted the architecture design in Part $A^2$. For KITTI detection evaluation, we directly load the learnt backbone for fine-tuning.

\section{Training Details}
\label{supp_sec:train_detail}

\subsection{Pretraining Details}
\label{supp_sec:pretrain}
As mentioned in the main paper, we use a standard SGD optimizer with
momentum 0.9, cosine learning rate scheduler starting
from 0.12 to 0.00012 and train the model for 1000 epochs
with a batch size of 1024. We observed that pretraining for 400 epochs already gives good results, and 1000 epochs for pretraining only slightly improve the model.

\subsection{Experimental Details for \pointnetplus}
\label{supp_sec:pretrain_point}
For all the VoteNet fine-tuning evaluations, we use the original configurations~\cite{qi2019votenet}, where we apply Adam optimizer~\cite{KingmaB14} and use a base learning rate 0.001 with a 10$\times$ weight decrease at 80, 120 and 160 epochs. The model is trained for 180 epochs in total. Since the training set of \stanford~\cite{2D-3D-S} only contains 200 training instances, we train for 360 epochs. We use a batch size of 8 for \scannet, \matterport, and \stanford, and a batch size of 16 for \sunrgbd. We use the same configuration for training from scratch and fine-tuning, and we only load the pretrained \pointnetplus backbone during fine-tuning. 

For the H3DNet fine-tuning, we only use one backbone network instead of the original four~\cite{zhang2020h3dnet}. For initial learning rate and decays, we use the original configurations. We found that with $3\times$ \pointnetplus backbone, we are able to re-produce the previous results reported in the paper with one backbone network. We use the same configuration for training from scratch and fine-tuning, and we only load the pretrained \pointnetplus backbone during fine-tuning.

\begin{table}[!b]
	\centering
	\setlength{\tabcolsep}{0.2em}\resizebox{\linewidth}{!}{
		\begin{tabular}{@{}l|cc@{}}
			\toprule
			\textbf{Task} & Scratch & \algorithmName \\
			\hline
			\textbf{\modelnet Linear} (Accuracy)& 50.7 & \textbf{85.4}\\
			\textbf{\modelnet Finetune} (Accuracy)& 88.2 & \textbf{91.3}\\
			\bottomrule
		\end{tabular}
	}
	\caption{\textbf{\modelnet classification}. We evaluate models by linear probing and full fine-tuning. We pretrain \pointnetplus models using \algorithmName on the \scanVideo dataset.}
	\label{tab:supp:modelnet}
\end{table}

\subsubsection{Modelnet Classification}
\label{supp_sec:pretrain_model}
In~\cref{sec:analysis} of the main paper, we used the \modelnet dataset for transfer learning. We trained linear classifiers on fixed features for this task and measured the classification accuracy.
\vspace{0.05in}
\par \noindent \textbf{Full finetuning.} We now show results on the same task but using finetuning, \ie, all the parameters of the backbone model are updated.
We use the \pointnetplus backbone to extract per-point feature and apply max-pooling to get the final global feature vector. We then apply one linear layer to get the final class labels. We use SGD+momentum optimizer with an initial learning rate 0.01. We use multistep LR scheduler with  10$\times$ weight decrease at 8, 16 and 24 epochs. We train for 28 epochs. We apply the data augmentation used in~\cite{qi2019votenet} during training. 
We use the same configuration for training from scratch and fine-tuning, and we only load the pretrained \pointnetplus backbone during fine-tuning. For linear probing, we fix the pretrained weight and only fine-tune the last linear layer. In~\cref{tab:supp:modelnet}, we compare the fine-tuning and train from scratch results for both linear probing and full fine-tuning. The pretrained \pointnetplus model provides consistent improvements.

\subsubsection{Label efficiency of \pointnetplus}
\begin{figure}[!t]
	\centering
	\includegraphics[width=0.23\textwidth]{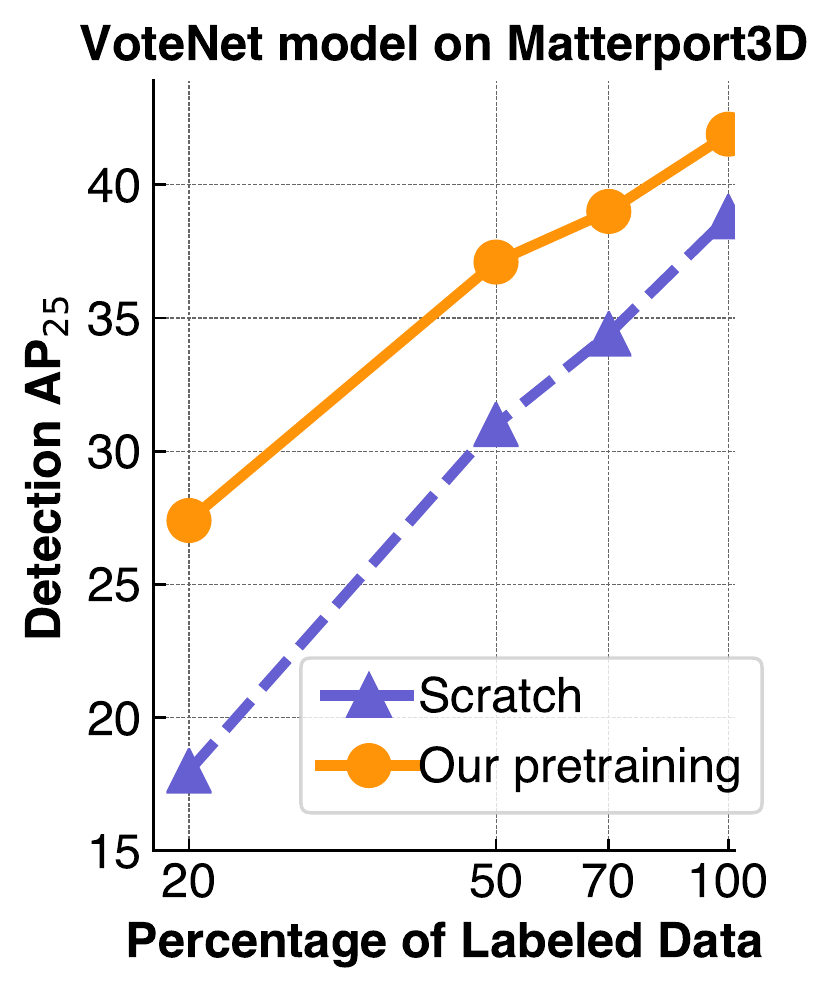}
	\includegraphics[width=0.237\textwidth]{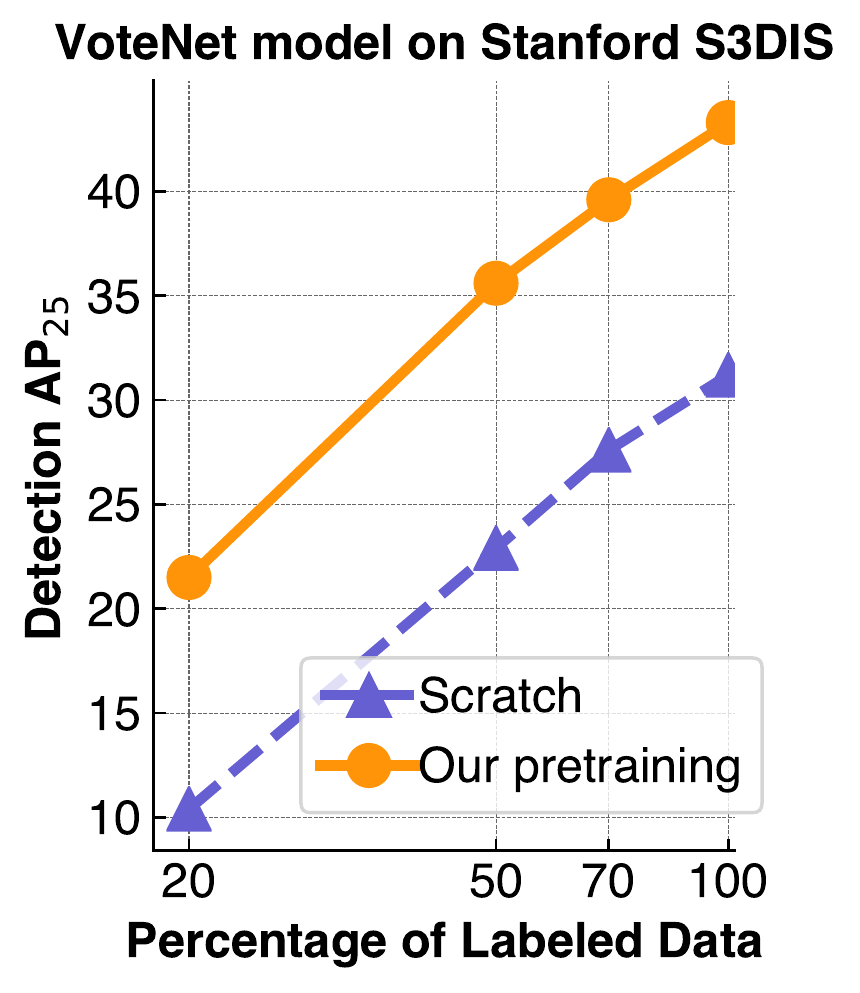}
	\caption{\textbf{Label-efficiency for \matterport and \stanford detection tasks.} We pretrain a \pointnetplus backbone model using our \algorithmName method on the \scanVideo dataset.}
	\label{fig:mp3d_stanford}
\end{figure}
We follow the same settings as~\cref{sec:label_efficiency} of the main paper and evaluate the label efficiency of the \pointnetplus pretrained using \algorithmName. In Figure 1 (main paper) we showed that \algorithmName pretraining was label efficient on the \scannet and \sunrgbd datasets. 
In~\cref{fig:mp3d_stanford}, we show the label efficiency plots for \matterport and \stanford downstream detection tasks. Our results are consistently better than training from scratch across all the detection benchmarks used in the paper.

\subsection{Experimental Details for \mink}
\label{supp_sec:pretrain_unet}

For all the \mink fine-tuning evaluation, we use SGD+momentum optimizer with an initial learning rate 0.1. We use Polynomial
LR scheduler with a power factor of 0.9. Weight decay is 0.0001. For voxel size, we use 0.05(5cm) for \stanford and \synthia and 0.04(4cm) for \scannet. We use the original data augmentation techniques in~\cite{choy20194d}. We use batch size 48 for \stanford and 56 for \scannet and \synthia.
We train the model with 8 V100 GPUs with data parallelism for 20000 iterations for all three tasks.
We use the same configuration for training from scratch and fine-tuning, and we only load the pretrained \mink backbone during fine-tuning. 

\subsubsection{\scannet Segmentation}
\begin{table}[t!]
	\centering
	\setlength{\tabcolsep}{0.2em}\resizebox{\linewidth}{!}{
		\begin{tabular}{@{}l|cc@{}}
			\toprule
			\textbf{Task} & Scratch & \algorithmName \\
			\hline
			\textbf{\scannet segmentation} (mIOU)& 70.3 & \textbf{71.2}\\
			\bottomrule
		\end{tabular}
	}
	\caption{\textbf{\scannet scene segmentation.} We evaluate \mink models after finetuning on the \scannet scene segmentation task. We pretrain the \algorithmName \mink models on the \scanVideo dataset.}
	\label{tab:supp:scannet}
\end{table}
For \scannet scene segmentation, due to memory issue, we increased the voxel size from the default 0.02(2cm) to 0.04(4cm), which leads to different scratch training results compared to~\cite{choy20194d}. Although we are pretraining and fine-tuning on the same dataset, our approach still provides improvements as shown in~\cref{tab:supp:scannet}.

\subsubsection{Label efficiency of \mink}
\begin{figure}[!b]
	\centering
	\includegraphics[width=0.4\textwidth]{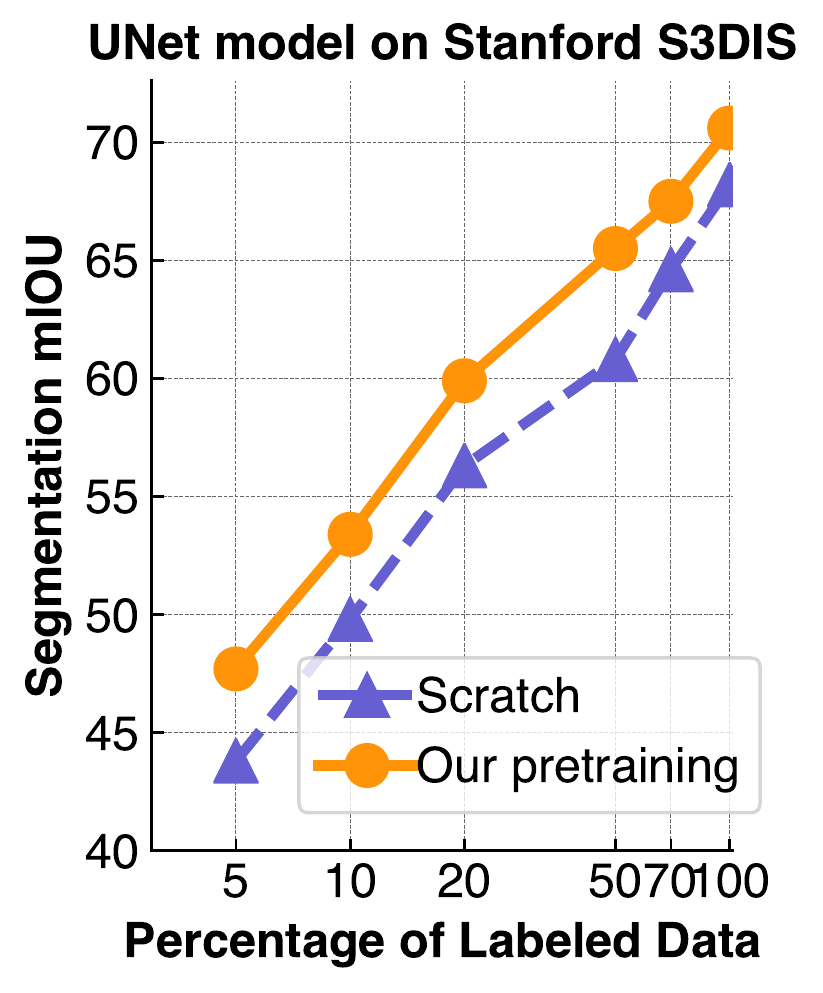}
	\caption{\textbf{Label-efficiency for \stanford scene segmentation} task using the \mink model on voxel input. \mink model was pretrained on \scanVideo using our \algorithmName method.}
	\label{fig:stanford_seg}
\end{figure}
We pretrain a \mink model using \algorithmName on the \scanVideo dataset. We finetune this model for scene segmentation task. In~\cref{fig:stanford_seg}, we show the data efficiency plot for \stanford scene segmentation dataset. Our approach provides consistent performance boost for different percentages of training data used.

\section{Experimental Details for Lidar Data}
\label{supp_sec:lidar_detail}
We now present the experimental details when using \algorithmName on LiDAR 3D data (Section 5.3 of the main paper). 

\begin{figure}[!b]
	\centering
	\includegraphics[width=0.23\textwidth]{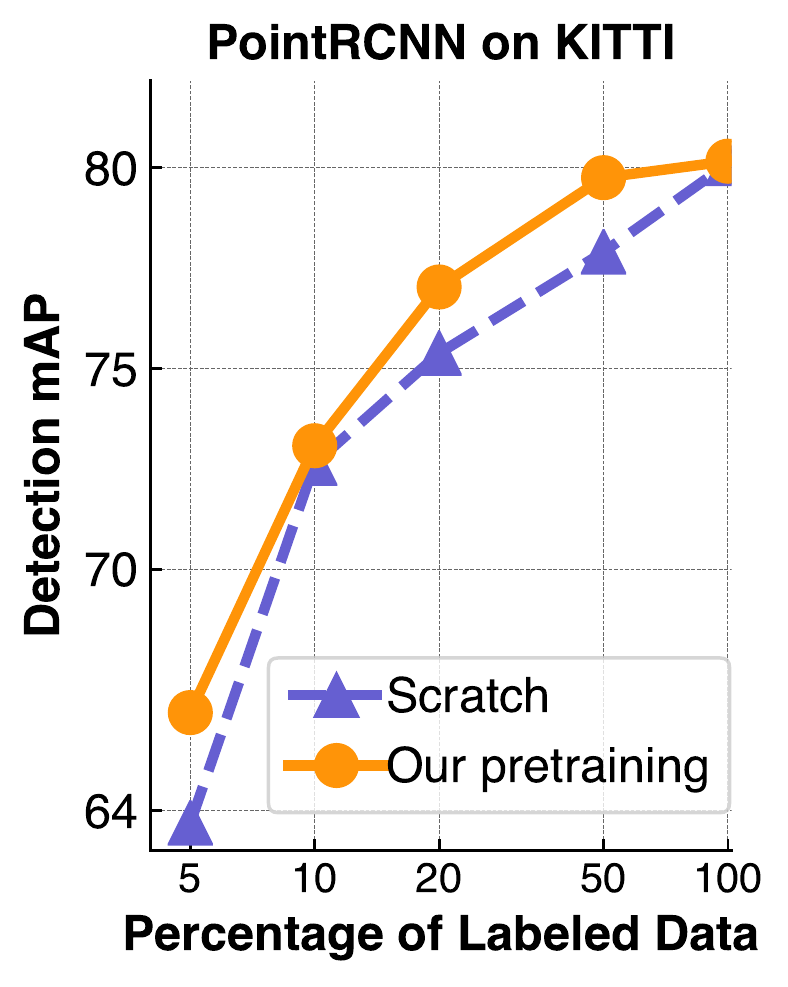}
	\includegraphics[width=0.23\textwidth]{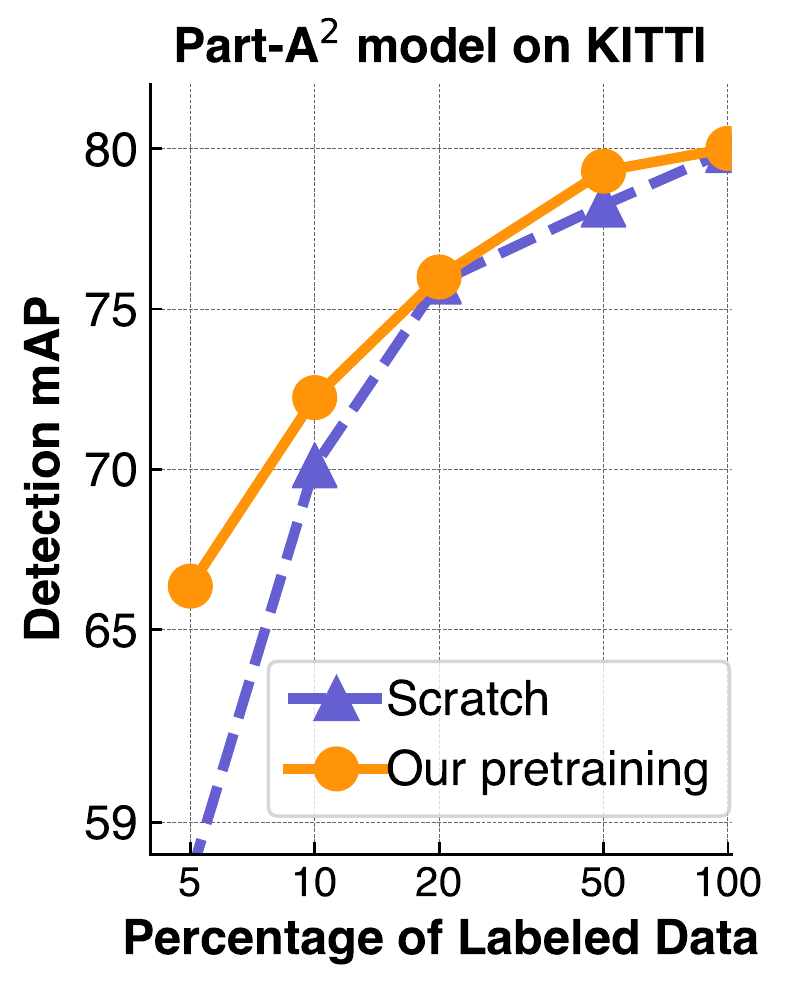}
	\caption{\textbf{Label-efficiency for KITTI \texttt{car}} detection at moderate difficulty level. We evaluate on the val split of the KITTI dataset.}
	\label{fig:lidar_car}
\end{figure}

\subsection{Fine-tuning Details}
\label{supp_sec:lidar_finetune}
We use the original configuration from PointRCNN~\cite{shi2019pointrcnn} and Part $A^2$~\cite{shi2019points} to get the scratch training results for different splits of training dataset. For PointRCNN, we use AdamW optimizer~\cite{loshchilov2017decoupled} with an initial learning rate 0.01, weight decay 0.01, and momentum 0.9. For Part $A^2$, we also use AdamW optimizer~\cite{loshchilov2017decoupled} with an initial learning rate 0.003, weight decay 0.01, and momentum 0.9. We use batch size 24 for PointRCNN and batch size 16 for Part $A^2$. We train both models for 80 epochs, and the learning rate will drop by $10\times$ at 35 and 45 epochs.
For both methods, we apply the same data augmentation and processing pipelines in~\cite{openpcdet2020}.

We use the same configuration for training from scratch and fine-tuning for 5\%, 10\%, 20\% and 50\% of the labeled training data. 
For 100\% training data, we observed over-fitting issue for classes with fewer training instances, such as cyclist and pedestrian. Thus, we increased the initial learning rate by $2\times$. For PointRCNN, we also decreased the number of training epochs to 60 and set the first weight drop at 30 epochs.
After modifying those parameters, we observed that the performance of scratch training didn't change.

\begin{figure}[!t]
	\centering
	\includegraphics[width=0.23\textwidth]{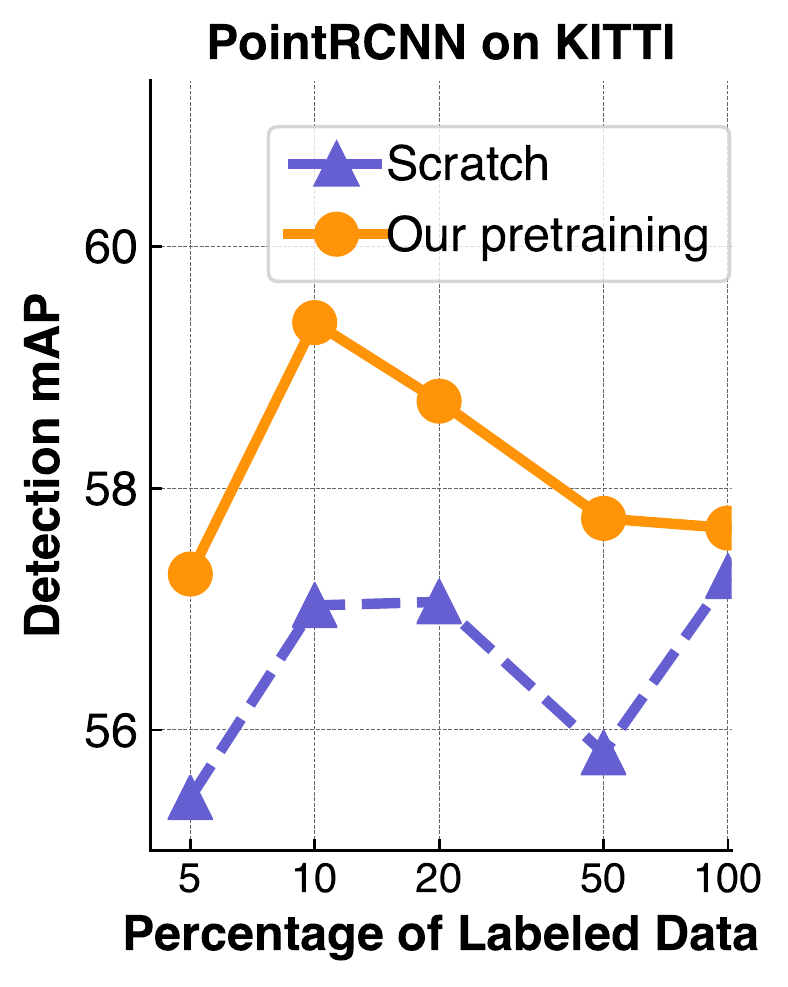}
	\includegraphics[width=0.23\textwidth]{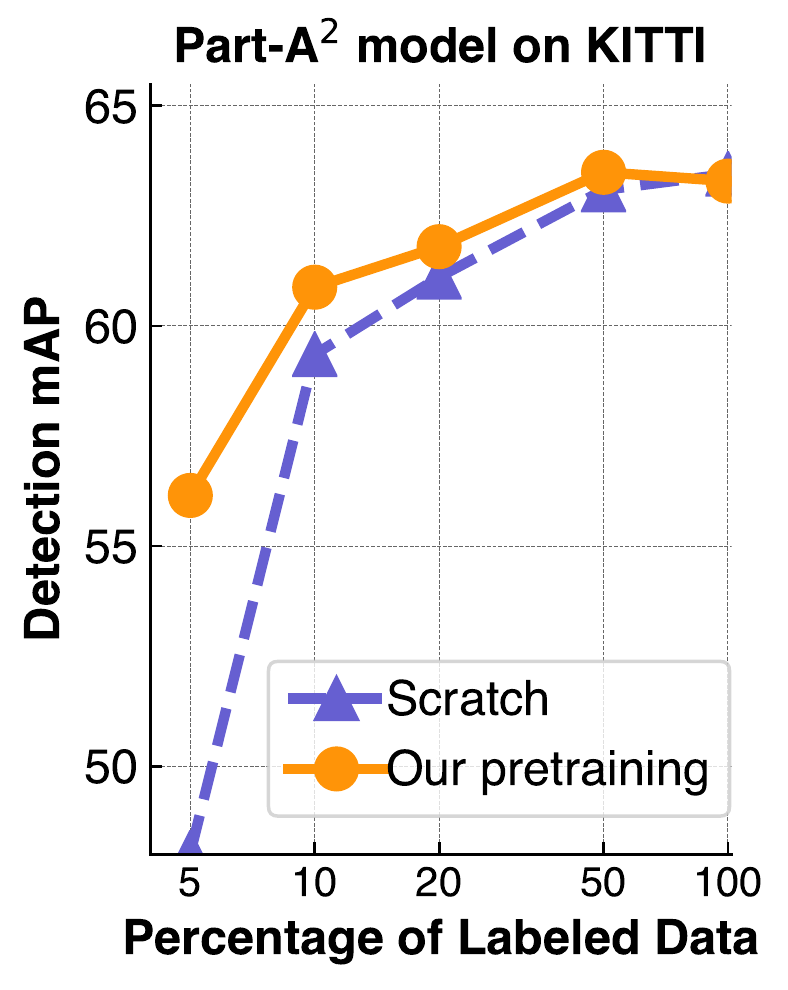}
	\caption{\textbf{Label-efficiency evaluation for KITTI \texttt{pedestrian} detection} at moderate difficulty level. We use the val split of the KITTI dataset.}
	\label{fig:lidar_ped}
\end{figure}

\begin{figure}
	\centering
	\includegraphics[width=0.23\textwidth]{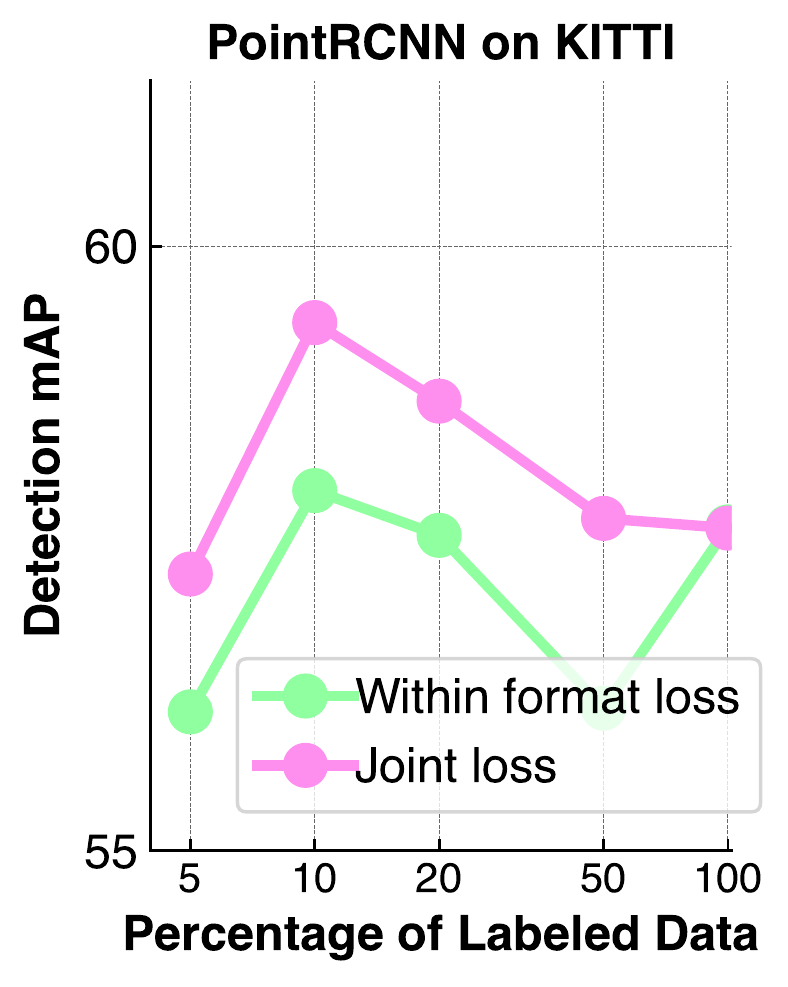}
	\includegraphics[width=0.23\textwidth]{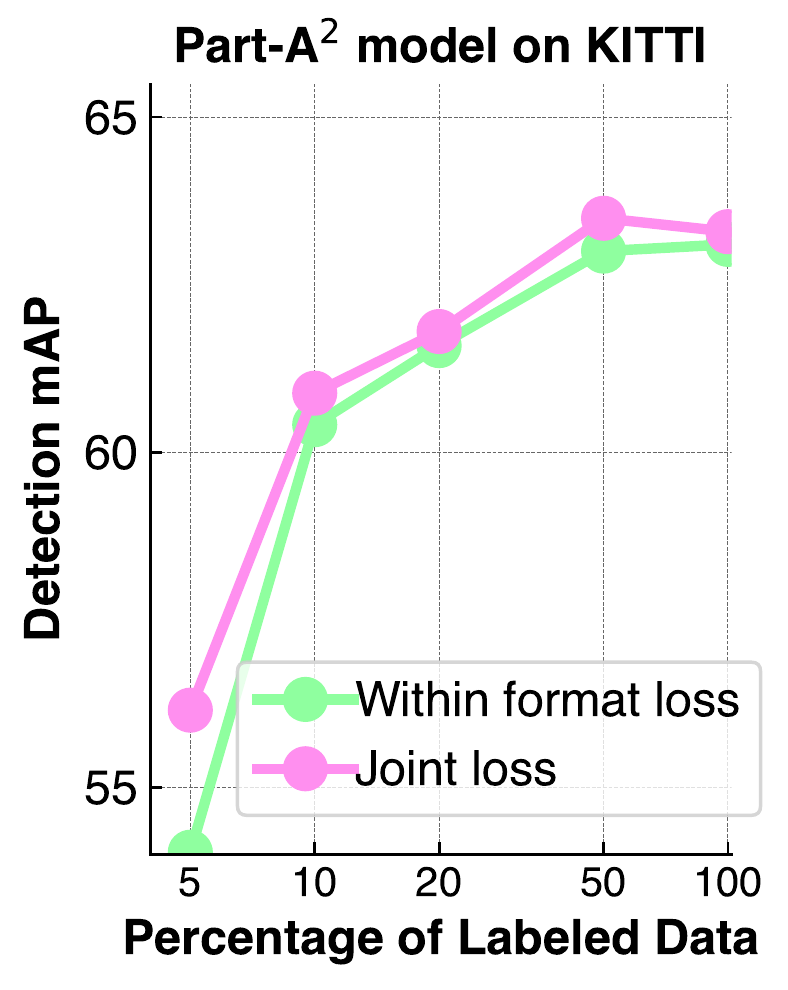}
	\caption{\textbf{Comparison between within format and joint training loss.} Label-efficiency evaluation for \texttt{pedestrian} detection at moderate difficulty level of the KITTI val split}
	\label{fig:lidar_ped_joint}
\end{figure}

\subsection{Results on KITTI}
\label{supp_sec:lidar_other}
In the main paper, due to space constraints, we only showed the results on the \texttt{cyclist} class in the KITTI dataset. We present results on the remainder classes.

We show the label efficiency evaluation results for \texttt{car} detection in KITTI in~\cref{fig:lidar_car}. For simplicity, we only show the results at moderate difficulty level. The results at other difficulty levels maintain similar pattern. \algorithmName provides a performance boost with fewer training instances, especially with 5\% of labeled data. 

In~\cref{fig:lidar_ped}, we show the label efficiency evaluation results for \texttt{pedestrian} detection in KITTI. Our pretraining provides consistent gain over scratch for PointRCNN model. For Part $A^2$, our pretraining also provides significant performance boost with fewer training instances.

\par \noindent \textbf{Advantage of joint training over within-format loss.} 
Similar to Section 4.3 of the main paper, we analyze if training jointly with the within and across-format losses provides better transfer performance. Specifically, we compare (Equation 3 of the main paper) our full joint loss with the within-format only loss. We pretrain separate models with these losses and evaluate them on the KITTI dataset.

In~\cref{fig:lidar_ped_joint}, we show the label efficiency evaluation results for \texttt{pedestrian} detection. We can see that with joint loss training, the model provides more performance gain with fewer training instances, which proves that the benefit of joint training holds across different pretraining data and architectures.

\end{document}